\documentclass[onefignum,onetabnum]{siamart190516}

\usepackage{textcomp, multirow, array, bigdelim, makecell, booktabs}
\usepackage{cite}
\usepackage[utf8]{inputenc}
\usepackage{amsmath,amssymb} %
\usepackage{graphicx} %
\usepackage{xcolor}
\usepackage{float}
\usepackage[left=1in, right=1in]{geometry}
\usepackage{undertilde}
\usepackage[utf8]{inputenc}
\usepackage{amsmath} %
\usepackage{graphicx} %
\usepackage{xcolor}
\usepackage{float}
\usepackage{gensymb}
\usepackage[left=1in, right=1in]{geometry}
\usepackage{multirow}
\usepackage{mathtools}
\usepackage{ mathrsfs }
\usepackage{kbordermatrix}%

\usepackage{enumitem}
\usepackage{subfigure}
\newlist{todolist}{itemize}{2}
\setlist[todolist]{label=$\square$}
\usepackage{pifont}
\usepackage{lipsum}
\usepackage{amsfonts}
\usepackage{epstopdf}
\usepackage{algorithmic}
\ifpdf
  \DeclareGraphicsExtensions{.eps,.pdf,.png,.jpg}
\else
  \DeclareGraphicsExtensions{.eps}
\fi

\newcommand{\pers}{\textrm{pers}}

\newcommand{\e}{\epsilon}

\newcommand{\boldA}{\mathbf{A}}

\newcommand{\R}{\mathbb{R}}

\newcommand{\Z}{\mathbb{Z}}

\renewcommand{\phi}{\varphi}
\renewcommand{\epsilon}{\varepsilon}

\newcommand\revision[1]{\textcolor{black}{#1}}

\usepackage{graphicx}
\graphicspath{{./figures/}{./Figures/}}

\newcommand{\myfiguresizeB}{0.32}

\newsiamremark{remark}{Remark}
\newsiamremark{hypothesis}{Hypothesis}
\crefname{hypothesis}{Hypothesis}{Hypotheses}
\newsiamthm{claim}{Claim}

\headers{Topological Signal Processing using the Weighted Ordinal Partition Network}{A.~Myers, F.~A.~Khasawneh, and E.~Munch}

\title{Topological Signal Processing using the Weighted Ordinal Partition Network%
\thanks{
Submitted to the editors DATE.
Associated python code: \href{https://lizliz.github.io/teaspoon/}{lizliz.github.io/teaspoon/}
\funding{This material is based upon work supported by the Air Force Office of Scientific Research under award number FA9550-22-1-0007.
}
}
}

\author{Audun Myers\thanks{Department of Mechanical Engineering, Michigan State University, East Lansing, MI (\email{myersau3@msu.edu}, \href{http://www.audunmyers.com}{audunmyers.com}).}
\and Firas A. Khasawneh\thanks{Department of Mechanical Engineering, Michigan State University, East Lansing, MI (\email{khasawn3@egr.msu.edu}, \href{https://www.firaskhasawneh.com}{firaskhasawneh.com}).}
\and Elizabeth Munch\thanks{Dept.~of Computation Mathematics Science and Engineering; and Dept.~of Mathematics, Michigan State University, East Lansing, MI (\email{muncheli@msu.edu}, \href{http://elizabethmunch.com}{elizabethmunch.com}).}
}

\usepackage{amsopn}

\ifpdf
\hypersetup{
  pdftitle={Topological Signal Processing using the Weighted Ordinal Partition Network},
  pdfauthor={A. Myers, F. A. Khasawneh, and E. Munch}
}
\fi

\begin{document}

\maketitle

\begin{abstract}
  \revision{
  One of the most important problems arising in time series analysis is that of bifurcation, or change point detection.
  That is, given a collection of time series over a varying parameter, when has the structure of the underlying dynamical system changed?
  For this task, we turn to the field of topological data analysis (TDA), which encodes information about the shape and structure of data.
  In this paper, we investigate a more recent method for encoding the structure of the attractor as a weighted graph, known as the ordinal partition network (OPN), representing information about when the dynamical system has passed between certain regions of state space.
  We provide methods to incorporate the weighting information, and show that this framework provides more resilience to noise or perturbations in the system as well as improving the accuracy of dynamic state detection.
  }
\end{abstract}



\section{Introduction} \label{sec:intro}

Time series are widely utilized to analyze dynamical systems with applications spanning everything from atmospheric science to zoology.
In many cases, the emphasis of the analysis is to determine when qualitative changes, known as bifurcations, have occurred in the underlying dynamical system.
This challenging task is important for predicting future response or to prevent detrimental system behavior.
For example, a change in measured biophysical signals can indicate upcoming health problems, and a change in the vibratory signals of machines or structures can be the harbinger of imminent failure.
Time series typically originate from real-life measurements of systems, and they provide only finitely sampled information from which the underlying dynamics must be gleaned.
This necessitates making conclusions on the continuous structure of dynamical systems using discretely sampled and often noisy time series.

Existing tools for detecting changes in the time series include
Lyapunov-based methods in the time domain
\cite{Tang1996,Froeschle1997,Guzzo2002,Lega2016},
frequency domain methods developed for Hamiltonian systems
\cite{Laskar1993,Sidlichovsky1996,Cordani2008},
entropy-based methods 
\cite{Nunez1996,Cincotta1999,Cincotta2000,Bandt2002,Myers2019},
Recurrence Plots (RPs)~\cite{Eckmann1987} which are related to 
the $\epsilon$-recurrence networks in graph theory 
\cite{Donner2010}, 
and the $0$-$1$ test for chaos and its extensions 
\cite{Gottwald2009,Tempelman2020,Tempelman2020a}.
However, Lyapunov exponents are difficult to estimate from time series, and their estimation is sensitive to noise and to a faithful reconstruction of the dynamics \cite{Parlitz2016}.
Frequency domain methods are predominantly applicable only for Hamiltonian systems, and they share with all the methods mentioned above the need for careful tuning of input parameters in order to extract useful dynamic state information.

\begin{figure}
    \centering
    \includegraphics[width = 0.9\textwidth]{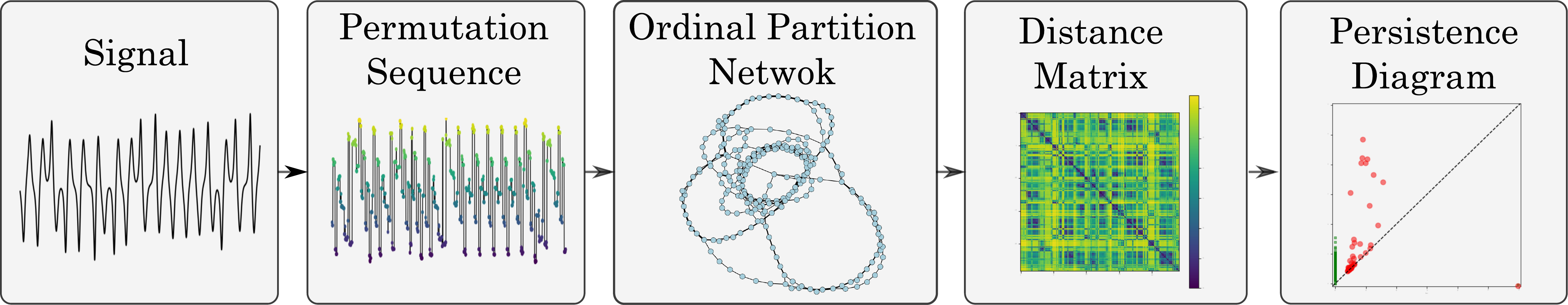}
    \caption{Pipeline for applying persistent homology to analyze the weighted transitional graph generated from a signal. In this work the state sequence is calculated as the permutation sequence.}
    \label{fig:pipeline}
\end{figure}

In this work, we utilize tools from the field of topological data analysis (TDA) \cite{Dey2021,Munch2017,Carlsson2009} to encode the shape and structure of the attractor of the underlying dynamical system in order to determine and analyze bifurcations in the system; see \cref{fig:pipeline}.
The idea of combining TDA with time series analysis is no longer new \cite{Perea2019a,Robinson2014}.
There are now many examples where persistent homology \cite{Oudot2015}, the flagship tool from TDA, can be used to measure the structure of a reconstructed attractor.
For example it can be used for
chatter detection in machining processes \cite{Khasawneh2016,Yesilli2019},
pulse counting in noisy systems \cite{Khasawneh2018b},
financial analysis \cite{Gidea2018a,Gidea2020},
periodicity detection \cite{Perea2014,Tempelman2020,Tempelman2020a} in video \cite{Tralie2017} and biological \cite{Perea2015} applications.

Persistent homology, colloquially referred to as persistence, encodes structure by analyzing the changing shape of a simplicial complex (a higher dimensional generalization of a network) over a filtration (a nested sequence of subcomplexes).
It should be noted that the majority of these applications utilize a relatively standard pipeline to construct this filtration.
Namely, given point cloud data embedded in $\R^n$ as input, construct the Vietoris Rips (VR) complex which includes a simplex at the maximum distance between any pair of its vertices.
Note that when applied to time series embeddings, this construction is closely related to the recurrence plots commonly used in the time series analysis literature \cite{Marwan2007a}.
The persistence diagram, which is a collection of points in $\R^2$ representing the appearance and disappearance of homological structures in the simplicial complex, can be computed and analyzed to determine whether two time series of interest have considerably distinct behavior.

Unfortunately, the persistence of point cloud data does have its drawbacks, in particular since VR complexes can become quite large (namely exponential) relative to the number of points in the original point cloud.
For this reason, recent work has begun to investigate alternative representations of an attractor in a way which captures information on its structure while remaining computationally reasonable.
To this end, we turn to network based representations of time series~\cite{Small2013}, focusing on the ordinal partition network~\cite{McCullough2015}.
Similar to the delay coordinate embedding, we study point clouds $\chi$ with points as
$X(t) = [x(t), x(t+\tau), \cdots, x(t + (n-1)\tau)] \in \chi$
but now map them to their permutation induced by the ordering; i.e.~the choice of permutation $\pi$ in the set of  $n!$ possible permutations
for which $x(t + \pi(0)\tau) \leq x(t+\pi(1)\tau) \leq \cdots \leq x(t+\pi(n-1)\tau)$.
We can then track the changing permutation as $t$ is varied in an ordinal partition network: each permutation $\pi$ becomes a vertex, and we include an edge from $\pi_i$ to $\pi_j$ if increasing $t$ passes from one permutation to the other. A more detailed introduction is provided in Section~\ref{sec_background}.

Our prior work~\cite{Myers2019} computed persistent homology of this construction to show that it could be used to differentiate between different kinds of dynamical system behavior; however, that work did not make use of a great deal of information available in the ordinal partition network when performing the analysis.
First, there is an inherent directionality on edges as we are always passing from one permutation to another; and second the number of times an edge is utilized can be viewed as weighting information on the network.
In this paper, we seek to make use of at least part of this additional information; namely we will incorporate the weighting information on the network and show that this provides better results, particularly in the case of dynamic state detection.

\subsection{Motivation and Our Contribution}
\begin{figure}
    \centering
    \includegraphics[width = 0.85\textwidth]{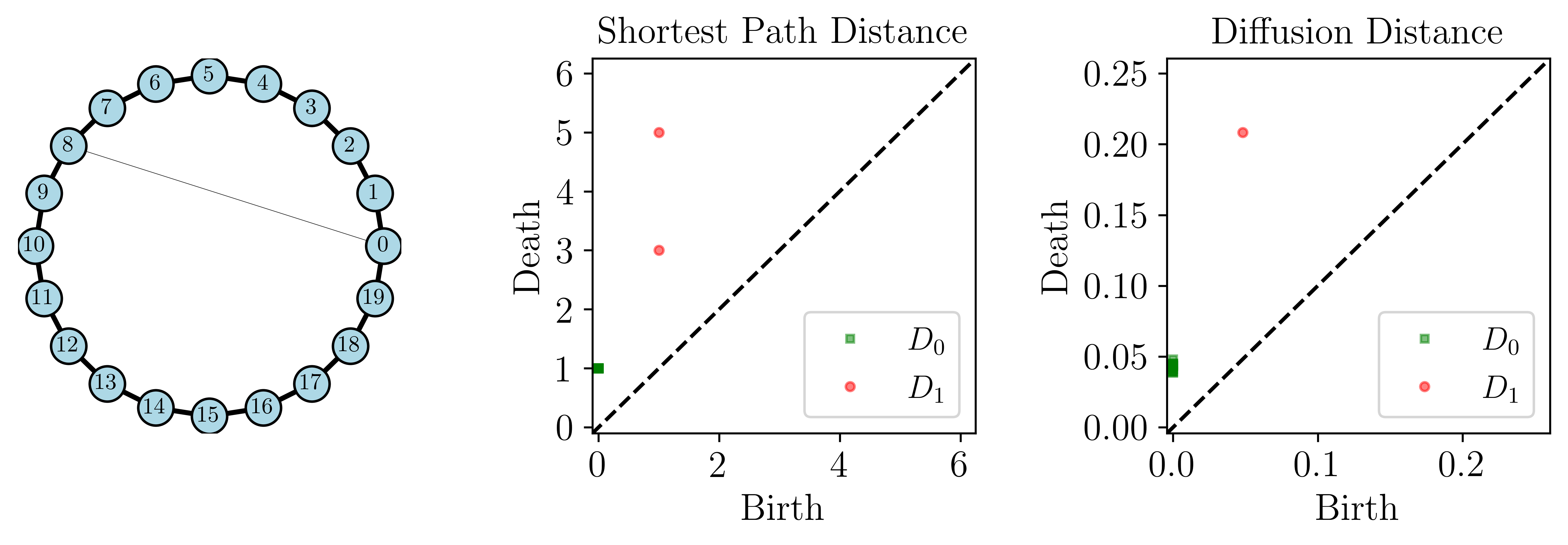}
    \caption{Example weighted cut cycle graph. The heavily weighted cycle is not easily seen in the persistence diagram computed from the unweighted shortest path distance (center), but is correctly identified when using the diffusion distance.}
    \label{fig:shortest_path_example_cut_cycle}
\end{figure}

The importance of considering weight information in graph representation of time series, such as those arising from the Ordinal Partition Network (OPN)~\cite{McCullough2015}, is demonstrated with a simple heavily weighted cycle graph with a low weighted cut edge as in the left of Fig.~\ref{fig:shortest_path_example_cut_cycle}.
The cut edge could be caused by additive noise, a perturbation to the system, or simply a falsely added state transition in the network formation process.
We would expect our measurements to find the large, heavily weighted cycle, shown as a single point far from the diagonal in the persistence diagram.
However, in this example, the shortest unweighted path distance considers all edges as equal, and thus the persistence diagram (center) has two off diagonal points for the two loops generated by the circle is split in two.
We correct this shortcoming by incorporating weight information (diffusion distance in this example), and the resulting diagram shown at right correctly identifies the heavily weighted circle as being more prominent.

The goal of this work is to investigate how to fruitfully incorporate weighted edges when calculating the persistent homology of OPNs. To do this, we investigate the use of the diffusion distance, the shortest weighted path distance, and the weighted shortest path distance between each node pair in the undirected and weighted network.
Using these distances we leverage topological data analysis for the characterization of ordinal partition networks.

\subsection{Organization}

The manuscript is organized as follows.
In Sec.~\ref{sec_background} we introduce state space reconstruction, OPNs, and persistent homology and how we apply it to complex networks.
In Sec.~\ref{sec_method} we provide details on the four distances (diffusion distance, shortest unweighted path distance, shortest weighted path distance, and weighted shortest path distance) and demonstrate how they are calculated for defining distances in networks and give empirical results for dynamic state detection in Sec.~\ref{sec_results}.
Additionally, in Sec.~\ref{ssec_example_two} we qualitatively show the differences in the resulting persistence diagrams for chaotic compared to periodic dynamics.
Following the initial example, in Sec.~\ref{sec_results} we demonstrate how the resulting persistence diagrams capture state changes in a signal using a nonlinear support vector machine kernel. Specifically, we show how the kernel can accurately separate the dynamic states using 23 example dynamical systems exhibiting both periodic and chaotic dynamics.
Additionally, in Sec.~\ref{sec_results}, we provide an empirical analysis of the additive noise robustness and stability of the resulting persistence diagrams in these example systems.

\section{Background}
\label{sec_background}
Before introducing our advancements to the method of analyzing complex weighted networks derived from dynamical system data, we first introduce the prerequisite background information.
\subsection{State Space Reconstruction and the Ordinal Partition Network}
\label{ssec:EmbeddingTimeSeries}

Takens' embedding theorem~\cite{Takens1981} allows us to use the technique of State Space Reconstruction (SSR) for the analysis of deterministic, nonlinear time series data from flows.
In summary, the theorem allows us to reconstruct a diffeomorphism of the original state space using only a single time series measurement from the dynamical system.
Specifically, the state of a dynamical system at time $t \in \mathbb{R}$ is defined as a vector $\mathbf{y} \in M \subseteq \R^n$, where $M$ is the manifold that the attractor lies on. The tracking of $\mathbf{y}$ over time is the flow $\phi^t(\mathbf{y})$.
However, our measurement of the underlying system is typically an observation function of the flow $\beta(\phi^t(\mathbf{x}))$.
Using an embedding time delay $\tau > 0$ and a sufficiently high embedding dimension $n \geq 2m+1$, where $m$ is the dimension of the manifold $M$, then the SSR will be (with high probability) diffeomorphic to the original attractor and can be used to study the dynamical system without loss of information.
For brevity we will call our observation function $x$ and then define the SSR vector $X(t)$ as
\begin{equation}
X(t) = [x(t), x(t+\tau), x(t + 2\tau), \ldots, x(t+(n-1)\tau].
\end{equation}

\begin{figure}
    \centering
    \includegraphics[scale = 0.9]{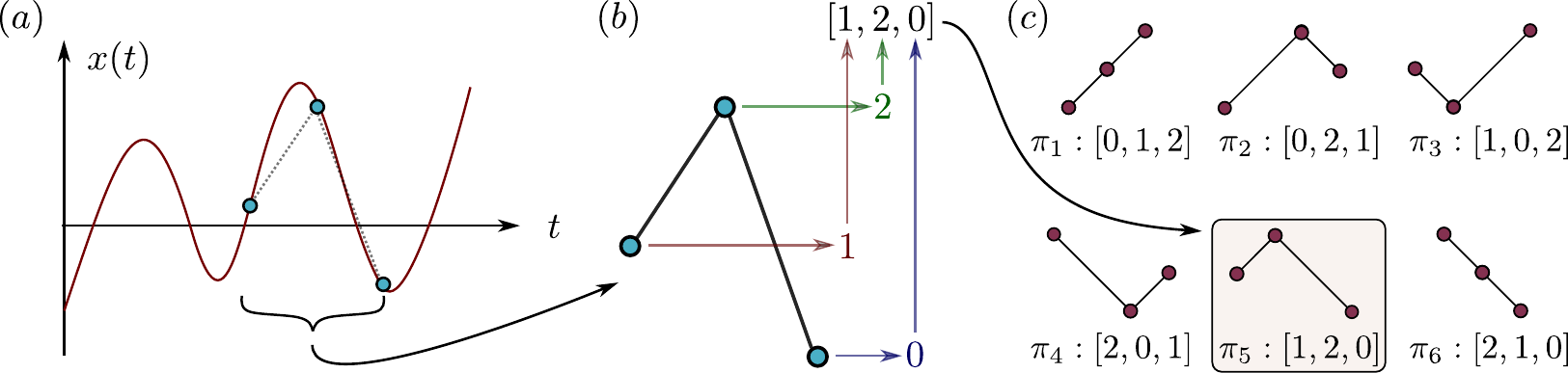}
    \caption{Ordinal partition state assignment example of SSR vector $\mathbf{x}_i \in \mathbb{R}^3$ in (a) to an ordinal partition in (b) and symbolic representation from 6 possible permutations for dimension $n=3$ in (c).}
    \label{fig:OP_symbol_example}
\end{figure}

In this paper, we forgo Takens embedding in favor of the ordinal partition network or graph representation of the state space.
A graph $G = (V,E)$ is a collection of vertices $V$ with edges $E = \{uv\} \subseteq V \times V$.
In this work all graphs are simple (no self-loops or multiedges) and undirected.
Additional stored information comes as a weighted graph, $G = (V,E, \omega)$ where $\omega:E \to \R_{\geq0}$ gives a non-negative weight for each edge in the graph.
Given an ordering of the vertices $V = \{v_1,\cdots, v_n\}$, a graph can be stored in an adjacency matrix $\boldA$ where the weighting information is obtained by setting $\boldA_{ij} = w_{(v_i, v_j)}$ if $v_iv_j \in E$ and 0 otherwise.
We also make comparisons to the unweighted graph where $\boldA_{ij} = \boldA_{ji} = 1$ if $v_iv_j \in E$ and 0 otherwise.

\begin{figure}
    \centering
    \includegraphics[scale = 0.9]{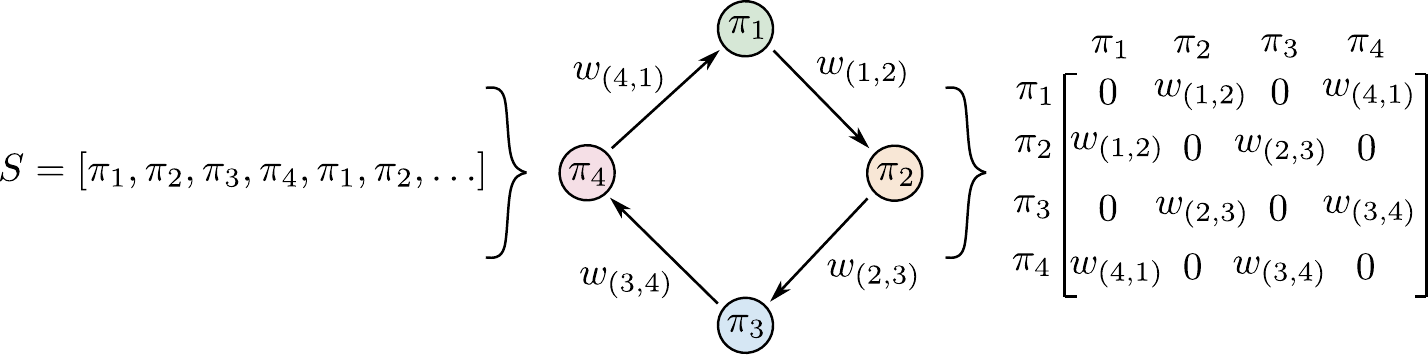}
    \caption{Example demonstrating formation of weighted and direct OPN and corresponding {(undirected)} adjacency matrix for a cyclic sequence of permutations as $S$.}
    \label{fig:weighted_OPN_example}
\end{figure}

The ordinal partition network \cite{Small2013,McCullough2015} provides a relatively simple method to assign symbolic representations for the SSR vectors to form a transition network.
This construction arose as a generalization of the concept of permutation entropy \cite{Bandt2002}.
Assume our data is provided as discretely sampled time series data $x = [x_1, x_2, \ldots, x_L]$ with $L$ as the number of samples from a signal sampled at uniform time stamps $t = [t_1, t_2, \ldots, t_L]$ with sampling frequency $f_s$.
An SSR vector of a discrete sampled signal is defined as
$X_i = [x_i, x_{i+\tau}, x_{i+2\tau}, \ldots, x_{i+\tau (n-1)}]$
with $i \in \mathbb{Z} \cap [1, L - \tau (n-1)]$, $\tau \in \mathbb{Z}$.
The basic idea of the OPN construction is to replace each SSR vector $X_i $ with a permutation $\pi$  where the vector $X_i$ is assigned to a permutation based on the sorted order of its coordinates.
Specifically, the permutation $\pi$ is the one in the set of $n!$ possible permutations
for which $x(t + \pi(0)\tau) \leq x(t+\pi(1)\tau) \leq \cdots \leq x(t+\pi(n-1)\tau)$, where $\pi(i)$ is the permutation value at index $i$; see Fig.~\ref{fig:OP_symbol_example} for an example.
Then the OPN is built with a vertex set of encountered permutations in the sequence $S$ with an edge included if the ordered point cloud passes from one permutation to the other.
An example for the case of a cyclic sequence of permutations can be seen in Fig.~\ref{fig:weighted_OPN_example}.

Note that the set of all permutations of dimension $n$ gives a cover of $\R^n$ with permutation $\pi_i$ representing a subspace of $\R^n$ given by the intersection of $\binom{d}{2}$ inequalities, and an edge is included based on passing from one of these subspaces to the other in one time step.
This partitioned symbolic representation of the state space allows the resulting ordinal partition network to capture meaningful topological information about the dynamical system's flow.

\subsection{Distances from a weighted graph}
\label{ssec:distances_from_graphs}
We next look at four different ways to define a distance between pairs of vertices given an input (weighted) graph.
In each case, we generate a distance matrix  $\mathbf{D}$ where entry $\mathbf{D}(a,b)$ gives the associated distance between vertices $a$ and $b$.

\begin{figure}
    \centering
    \includegraphics[width = 0.53\textwidth]{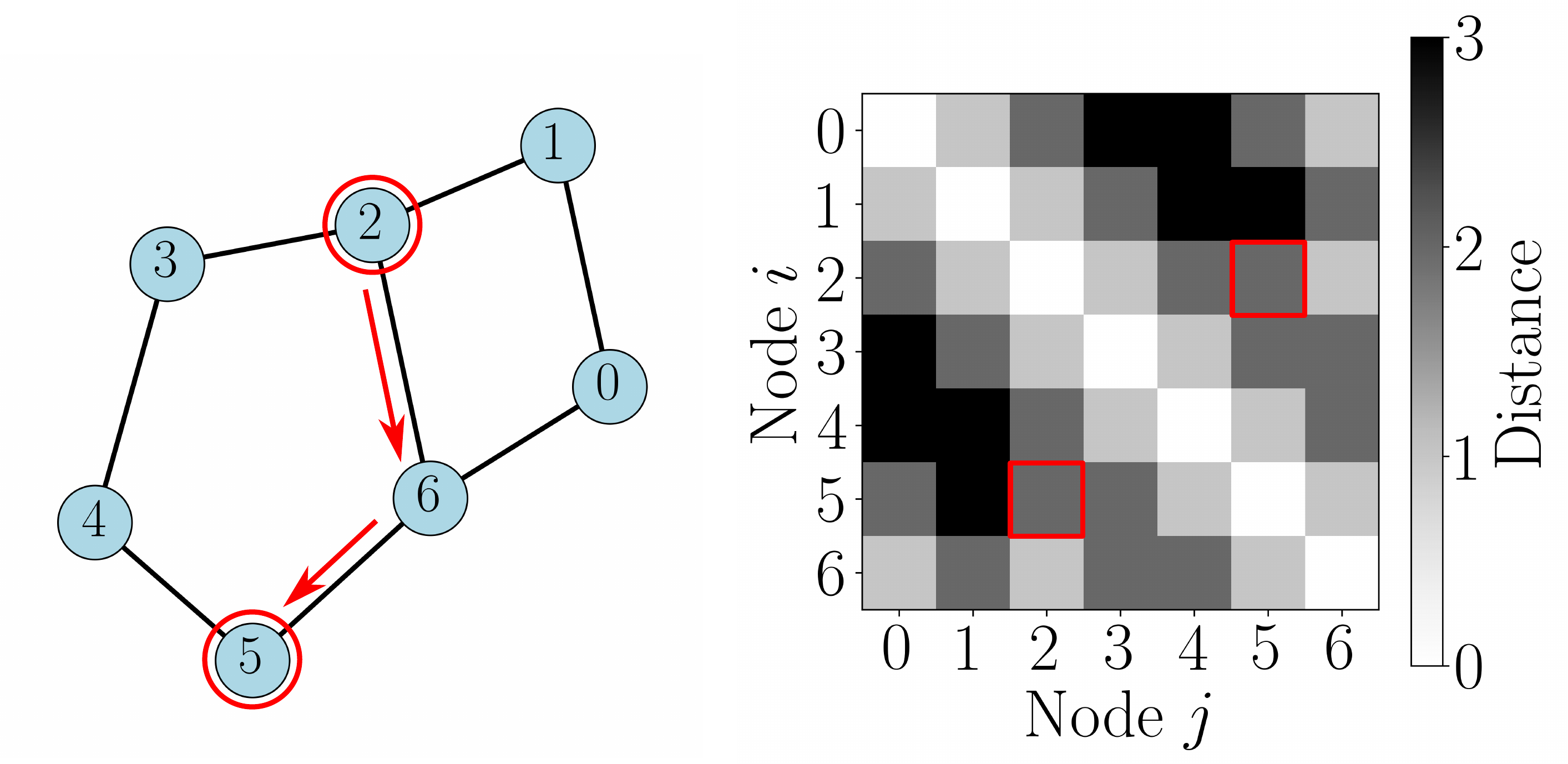}
    \caption{Example basic graph with corresponding shortest path distance matrix. Highlighted in red is an example shortest path from node $2$ to $5$ with shortest path distance $2$.}
    \label{fig:example_shortest_path}
\end{figure}

The first method, the shortest unweighted path distance, ignores the weighting information entirely, using only the number of edges to get from vertex $a$ to vertex $b$.
Specifically, $\mathbf{D}(a,b)$ is the number of steps it takes to transition from $a$ to $b$ through the shortest path.
See the example of Fig.~\ref{fig:example_shortest_path}.
The shortest path distance is calculated using the \texttt{NetworkX} implementation of Dijkstra's algorithm~\cite{Dijkstra1959} with the unweighted adjacency matrix.

The second method, the shortest weighted path, similarly only uses the number of edges between vertex $a$ and $b$ as the path distance. However, the weighted information is incorporated through the choice of the path. This is done by choosing the path with the lowest summed weight of all paths between $a$ and $b$. To make it such that the path with the largest weights is used, the inverse of the edge weights is used when calculating the shortest path. Again, this distance is calculated using the \texttt{NetworkX} implementation of Dijkstra's algorithm~\cite{Dijkstra1959} but with the inverse of the weighted adjacency matrix.

The third method, the weighted shortest path is very similar to the second method. The only variation is that the sum of the edge weights along the path is used as the distance. The path used is found using the inverse of the edge weights similar to the second method.

The fourth method for computing distances is the diffusion distance; for more details we direct the reader to~\cite{Coifman2006}.
This is computed using the transition probability distribution matrix  $\mathbf{P}$ of the graph, where $\mathbf{P}(a,b)$ is the probability of transitioning to vertex $b$ in the next step given you are currently at $a$.
Given the weighted, undirected adjacency matrix $\mathbf{A}$, the transitional probability matrix is calculated  as
\begin{equation*}
\mathbf{P}(i,j) = \frac{\mathbf{A}{(i,j)}}{\sum_{k=1}^{|V|} \mathbf{A}{(i,k)}}.
\end{equation*}
This formulation of the probability matrix only has transition probabilities greater then zero for one step neighbors of $i$.
However, the transition probabilities for non-adjacent neighbors of node $i$ can be calculated using the random walk and the diffusion process.
A random walk is the sequences of nodes visited $(a_1, a_2, \ldots)$ in $t$ steps, where the selection of the next node is based on the transition probabilities.
It is a classic exercise to show that, given $\mathbf{P}$, the probability distribution for transitioning to vertex $b$ from vertex $a$ in $t$ random walk steps is $\mathbf{P}^t(a, b)$.

The diffusion distance is a measure of the degree of connectivity of two nodes in a connected graph after $t$ steps using the lazy transition probability $\tilde{\mathbf{P}}^t$ based on the possible random walks of length $t$ and is calculated as
\begin{equation}
	d_t(a,b) = \sqrt{ \sum_{c \in V}  \frac{1}{\mathbf{d}(c)} { \left[ \tilde{\mathbf{P}}^t (a,c) - \tilde{\mathbf{P}}^t (b,c) \right] }^2 }
\end{equation}
where $\mathbf{d}$ is the degree vector of the graph with $\mathbf{d}(i)$ as the degree of node $i$ and $\tilde{\mathbf{P}}$ is the lazy transition probability matrix, where the initial zero diagonal of $P$ is set such that $\tilde{\mathbf{P}} = 1/2 (\mathbf{I} + \mathbf{P})$. In other words, there is an equal probability of staying and leaving at node $i$ in a single step. Applying the diffusion distance to all node pairs results in the distance matrix $\mathbf{D}_t$.

Consider the diffusion distance with two nodes having a connected path with high transition probability edges or many random walk paths connecting the two, then the diffusion distance between them will be low.
However, if two vertices are only connected through a single, low probability edge transition from a possible perturbation in the graph, then their diffusion distance will be large.
A common example implementing the diffusion distance is based on assigning $\mathbf{P}$ as a function of the proximity of nodes.
Using this formulation of the transition probability, it is possible to cluster the data based on the distances as demonstrated in~\cite{Coifman2006}.
However, due to the natural transitions that occur in transitional complex networks, the diffusion distance is a natural solution for incorporating edge weight data into the distance measurement.

\subsection{Persistent Homology} \label{ssec:persistent_homology}
In order to analyze the shape of the constructed graphs, we turn to a generalization of the graph known as a \textit{simplicial complex}, and a measurement tool known as \textit{persistent homology}.
We direct the interested reader looking for a more in depth discussion to \cite{Hatcher,Munkres2,Oudot2015,Dey2021}.

\paragraph{Simplicial complexes}

A simplicial complex is one way to  generalize the concept of a graph to higher dimensions.
Like a graph, we start with a (finite) vertex set $V$; a simplex $\sigma \subseteq V$ is any subset of vertices.
The dimension of a simplex $\sigma$ is $\dim(\sigma) = |\sigma|-1$.
Note that graph edges are thus simplices of dimension 1.
The simplex $\sigma$ is a face of $\tau$, denoted $\sigma \preceq \tau$, if $\sigma \subseteq \tau$.
A simplicial complex $K$ is a collection of simplices which is closed under the face relation; i.e.~if $\sigma \in K$ and $\tau \preceq \sigma$, then $\tau \in K$.
The dimension of a simplicial complex is the largest dimension of its simplices, $\dim(K) = \max _{\sigma \in K} \dim(\sigma)$.
The $d$-skeleton of a simplicial complex is all simplices of $K$ with dimension at most $d$, $K^{(d)} = \{ \sigma \in K \mid \dim(\sigma) \leq d\}$.

One way to build a simplicial complex from a graph input is to start with the graph as the 1-skeleton, and then include all higher dimensional simplices when possible:
\begin{equation}
  \label{eq:cliquecomplex}
 K(G) = \{ \sigma \subseteq V \mid uv \in E \text{ for all } u\neq v \in \sigma\}.
\end{equation}
This is called the clique complex.
The clique complex of the complete graph on $n$ vertices is called the complete simplicial complex on $n$ vertices.

\paragraph{Homology}

Traditional homology~\cite{Hatcher,Munkres2} counts the number of structures of a particular dimension in a given topological space, which in our context will be a simplicial complex.
In this context, the structures measured can be connected components (0-dimensional structure), loops (1-dimensional structure), voids (2-dimensional structure), and higher dimensional analogues as needed.
In this work, we focus on 0- and 1-dimensional homology.

Given a simplicial complex $K$, denote the $d$-dimensional simplices by $\sigma_1,\cdots, \sigma_\ell$.
A $d$-dimensional chain is a formal sum of the $d$-dimensional simplices $\alpha = \sum_{i=1}^\ell a_i \sigma_i$.
We assume the coefficients $a_i \in \Z_2 = \{0,1\}$ and addition is performed mod 2; i.e.~$1+1=0$.
For two chains $\alpha = \sum_{i=1}^\ell a_i \sigma_i$ and $\beta = \sum_{i=1}^\ell b_i \sigma_i$, $\alpha + \beta = \sum_{i=1}^\ell (a_i + b_i) \sigma_i$.
The collection of all $d$-dimensional chains forms a vector space denoted $C_d(K)$ with addition given by addition of coefficients.
The boundary of a given $d$-simplex is
\begin{equation*}
 \partial_d(\sigma) = \sum_{ \substack{\tau \prec \sigma\\ \dim(\tau) = d-1}} \tau.
\end{equation*}
That is, the boundary is the formal sum of faces which are exactly one lower dimension.
If $\sigma$ is a vertex so that $\dim(\sigma) = 0$, we set $\partial_d(\sigma) = 0$.
Then the boundary operator $\partial_d:C_d(K) \to C_{d-1}(K)$ is given by
\begin{equation*}
\partial_d(\alpha) = \partial_d\left(\sum_{i=1}^\ell a_i \sigma_i \right)
 = \sum a_i \partial_d(\sigma_i).
\end{equation*}

A $d$-chain $\alpha \in C_d(K)$ is a cycle if $\partial_d(\alpha) = 0$; it is a boundary if there is a $d+1$-chain $\beta$ such that $\partial_{d+1}(\beta) = \alpha$.
The group of $d$-dimensional cycles is denoted $Z_d(K)$; the boundaries are denoted $B_d(K)$.
In particular, any $0$-chain is a  $0$-cycle since $\partial_0(\alpha) = 0$ for any $\alpha$.
A $1$-chain is a $1$-cycle iff the 1-simplices (i.e., edges) with a coefficient of 1 form a closed loop.
It is a fundamental exercise in homology to check that $\partial_{d} \partial_{d+1} = 0$ and therefore that $B_d(K) \subseteq Z_d(K)$.

The $d$-dimensional homology group is defined to be $H_d(K) = Z_d(K)/B_d(K)$.
An element of $H_d(K)$ is called a homology class and is denoted $[\alpha]$ for $\alpha \in Z_d(K)$ where $[\alpha] = \{ \alpha + \partial(\beta) \mid \beta \in C_{d+1}(K)\}$.
We call $\alpha$ a representative of $[\alpha]$, noting that any element of $[\alpha]$ can be used as its representative so this choice is by no means unique.
In the particular case of 0-dimensional homology, there is a unique class in $H_0(K)$ for each connected component of $K$.
For $1$-dimensional homology, there is one homology class in $H_1(K)$ for each ``hole'' in the complex.
\paragraph{Persistent homology}

\begin{figure}
    \centering
    \includegraphics[width = 0.99\textwidth]{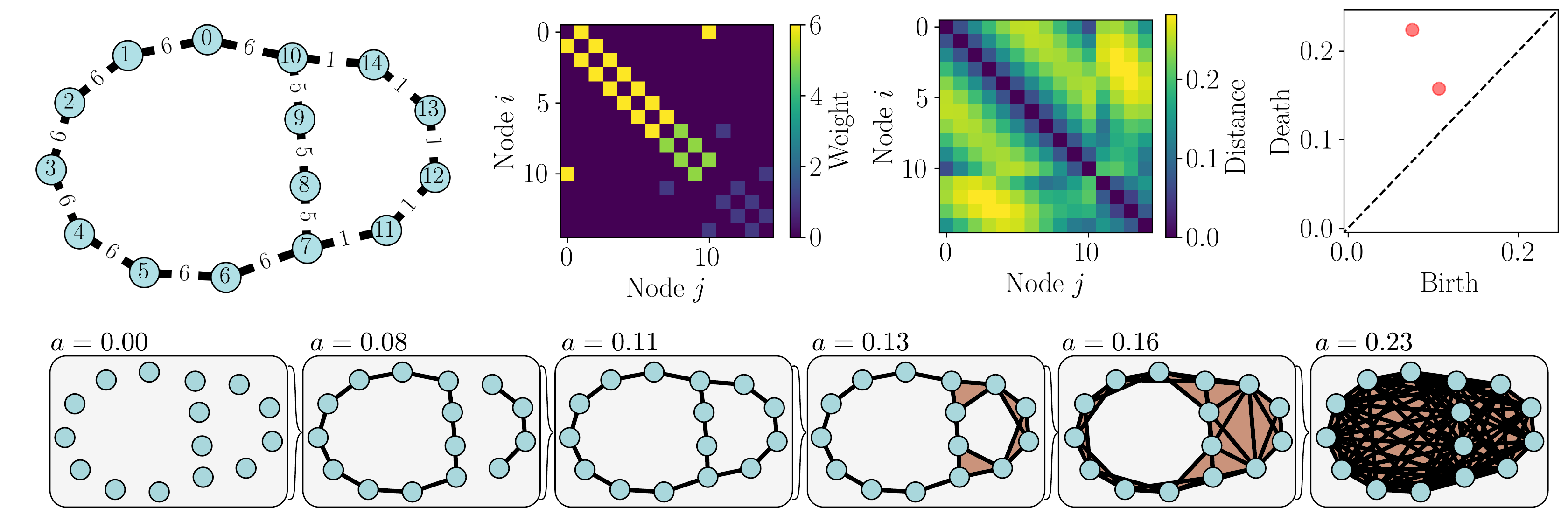}
    \caption{Persistent homology of weighted complex network. Top left shows the weighted network with corresponding adjacency matrix to its right.
    Third is the distance matrix and then at the top right is the persistence diagram of one-dimensional features. The bottom row shows the filtration at critical values.}
    \label{fig:persistent_homology_example}
\end{figure}

We next look to a more modern viewpoint of homology which is particularly useful for data analysis, persistent homology.
In this case, we study a \textit{changing} simplicial complex and encode this information via the \textit{changing} homology.
In explaining persistence, we will follow the example of Fig.~\ref{fig:persistent_homology_example} for the setting used in this work where the input data is a weighted network.

A filtration of a simplicial complex $K$ is a collection of nested simplicial complexes
\begin{equation*}
  K_1 \subseteq K_2 \subseteq \cdots \subseteq K_N = K.
\end{equation*}
See the bottom row of Fig.~\ref{fig:persistent_homology_example} for an example of a filtration.
In this work, we will be focused on the following filtration which arises from finite metric space; in our case, this is given as a pairwise distance matrix $\mathbf{D} \in \R_{\geq 0} ^{n \times n}$, obtained from a weighted graph as described in Sec.~\ref{ssec:distances_from_graphs}.
Set the vertex set to be $V = [1,\cdots, n]$ and for a fixed $a \in \R$, let
\begin{equation*}
  K_a = \{\sigma \subset V \mid \mathbf{D}(u,v) \leq a \text{ for all }u \neq v \in \sigma \}.
\end{equation*}
This can be thought of as the clique complex (Eq.~\refeq{eq:cliquecomplex}) on the graph with edges given by all pairs of vertices with distance at most $a$.
Further, since $K_a \subseteq K_b$ for $a \leq b$, this construction gives rise to a filtration
\begin{equation*}
  K_{a_1} \subseteq K_{a_2} \subseteq \cdots \subseteq K_{a_N}
\end{equation*}
for any collection $a_1 \leq a_2 \leq \cdots \leq a_N$.

Fix a dimension $d$.
For any inclusion of one simplicial complex to another $L \hookrightarrow K$, there is an induced  map on the $d$-chains $\iota: C_d(L) \to C_d(K)$ by simply viewing any chain in the small complex as one in the larger.
Less obviously, this extends to a map on homology $\iota_*:H_d(L) \to H_d(K)$ by sending $[\alpha] \in H_d(L)$ to the class in $H_d(K)$ with the same representative.
That this is well defined is a non-trivial exercise in the definitions~\cite{Hatcher}.
Putting this together, given a filtration
\begin{equation*}
K_{a_1} \subseteq K_{a_2} \subseteq \cdots \subseteq K_{a_N}
\end{equation*}
 there is a sequence of linear transformations on the homology
\begin{equation*}
  H_d(K_{a_1}) \to H_d(K_{a_2}) \to \cdots \to H_d(K_{a_N}).
\end{equation*}
A class $[\alpha] \in H_d(K_{a_i})$ is said to be born at $a_i$ if it is not in the image of the map $H_d(K_{a_{i-1}}) \to H_d(K_{a_i})$.
The same class dies at $a_j$ if $[\alpha] \neq 0$ in $H_d(K_{a_{j-1}})$ but $[\alpha] = 0$ in $H_d(K_{a_{j}})$.
In the case of 0-dimensional persistence, this feature is encoding the appearance of a new connected component at $K_{a_i}$ that was not there previously, and which merges with an older component entering $K_{a_j}$.
For 1-dimensional homology, this is the appearance of a loop structure that likewise fills in entering $K_{a_j}$.

The persistence diagram encodes this information as follows.
For each class that is born at $a_i$ and dies at $a_j$, the persistence diagram has a point in $\R^2$ at $(a_i,a_j)$.
Because several features can appear and disappear at the same times, we allow for repeated points at the same location.
For this reason, a persistence diagram is often denoted as a multiset of its off-diagonal points, $D = \{(b_1,d_1),\cdots,(b_k,d_k)\}$.
See the top right of Fig.~\ref{fig:persistent_homology_example} for an example.
Note that the farther a point is from the diagonal, the longer that class persisted in the filtration, which signifies large scale structure.
The \textit{lifetime} or \textit{persistence} of a point $x = (b,d)$ in the diagram in a persistence diagram $D$ is given by $\pers(x) = |b-d|$. It is often of interest to investigate only a specific subset of $d$ dimensional features from a persistence diagram, which we represent as $D_d$.

\subsection{\revision{Distances between Diagrams and Multi-dimensional Scaling (MDS)}}
\label{ssec:MDS}

\revision{
  In order to understand how similar two persistence diagrams are to each other, we turn to the bottleneck distance between diagrams. 
  The idea is to provide a number $d(D, D')$ which is small if the two diagrams $D$ and $D'$  are similar. 
  Since diagrams could have different numbers of points, we must be a bit careful as to how to define this. 
  Further, we want this distance to encode the idea that points far from the diagonal represent long lived features, while those close the diagonal are short lived features and thus the latter should be similar to a diagram which does not have these at all. 
}

\revision{  
  One of the most common options, the bottleneck distance, is given as follows. 
  Let the two diagrams in question be defined in terms of their off diagonal points, so  
  $D = \{(b_1,d_1),\cdots,(b_k,d_k)\}$ and  
  $D' = \{(b'_1,d'_1),\cdots,(b'_{k'},d'_k)\}$.
  A partial matching is defined as a matching on a subset of the points of the diagram, i.e.~for a pair $S \subseteq D$ and $S' \subseteq D'$, we have a bijection $\phi:S \to S'$. 
  Then the cost of the partial matching is given as the maximum of the $L_\infty$ distance between the matched points (those in $S$ and $S'$), and of the distance between the unmatched points  and the diagonal. 
  Specifically, 
  \begin{equation*}
    C(\phi) = \max\bigg\{ 
      \{ \|a_i - \phi(a_i)\|_\infty \}
      \cup 
      \{ \tfrac{1}{2} (d-b) \mid (b,d) \in D \setminus S \}
      \cup 
      \{ \tfrac{1}{2} (d'-b') \mid (b',d') \in D' \setminus S' \}
    \bigg\}.
  \end{equation*}
  The bottleneck distance is then defined to be 
  \begin{equation*}
    d_B(D,D') = \min_{\phi} C(\phi),
  \end{equation*}
  that is, the minimum cost over all possible partial matchings $\phi$.
}

\revision{
In this paper, we will be generating a persistence diagram to represent each input time series. 
In order to visualize the resulting separation in the metric space of persistence diagrams, we will perform a particular case of lower dimensional embedding, namely multidimensional scaling (MDS) to show separations. 
Given a finite metric space, in our case a collection of persistence diagrams $\{ D_i\}$,  with a choice of distance, in our case bottleneck distance $d_B$, MDS looks for a lower dimensional embedding of the data that preserves the distances as much as possible.
That is, we wish to find $X:= \{x_i\} \subseteq \R^d$ minimizing 
\begin{equation*}
  \mathrm{Stress}(X) = \left(\sum_{i \neq j} (d_B(D_i,D_j) - \|x_i-x_j\|)^2\right).
\end{equation*} 
It should be noted that a major drawback of MDS that it cannot be used for machine learning processes such as supervised learning.  
In particular, that would require being able to learn the embedding from the training data set, and then determining the location for an unseen test data point, however this is not possible with the MDS framework. 
For this reason, we use MDS in this paper as a visual inspection tool to determine whether the persistence diagrams for the labeled classes are separated in persistence diagram space, rather than for traditional machine learning proceedures.
}

\section{Method}  \label{sec_method}
In our previous work~\cite{Myers2019}, we investigated how the one-dimensional persistent homology of the shortest path distance for an unweighted complex network can be used to analyze the dynamics of the system.
A natural extension of this work is on how we can incorporate more information about the graph--such as the edge weight--to better measure the shape of the graph through persistent homology.
Figure~\ref{fig:pipeline} shows the pipeline investigated here and an example can be seen in Fig.~\ref{fig:persistent_homology_example}.

We begin with a signal and construct the weighted ordinal partition network as described in Sec.~\ref{ssec:EmbeddingTimeSeries}.
As in the case of Takens' embedding, we must take care when choosing the parameters $\tau$ and $n$.
Theoretically, nearly any time lag $\tau$ can be used for a completely noise-free and an infinitely precise signal, but in practice a $\tau$ is typically chosen using the method of mutual information~\cite{Fraser1986} or autocorrelation~\cite{Box2015a} for state space reconstruction. However, in this work we choose $\tau$ using the method of multi-scale permutation entropy as suggested in~\cite{Myers2020c} since we are forming permutations to construct the OPN.
While an appropriate embedding dimension $n$ for the state space reconstruction may be sufficient, it may not be a high enough dimension to capture the complexity of the time series.
To alleviate this issue, Bandt and Pompe~\cite{Bandt2002} suggested using higher dimensions (e.g. $n \in [4, 10]$) to allow for $n!$ different states to better capture the complexity of the time series. In this work we will use a dimension $n=6$ unless otherwise stated.

We next choose a distance for the given graph input as described in Sec.~\ref{ssec:distances_from_graphs}.
In order to compare with prior work, we examine the shortest unweighted path distance, while our main focus is on the weight-incorporating shortest path distances and the diffusion distance.
Given the distance matrix, we then compute the persistence diagram as described in Sec.~\ref{ssec:persistent_homology}.

It is important to mention the sensitivity of the diffusion distance $\mathbf{D}_t$ to the selection of the number of walk steps $t$.
We used an empirical study of 23 continuous dynamical systems to determine the optimal $t$ such that a periodic signal creates a significant point in the persistence diagram representing the cycle.
More details on this analysis are available in the appendix in Section~\ref{appendix_section_on_t}.
We found an optimal value of $d<t<3d$, where $d$ is the diameter of the graph. Specifically, the diameter is measured as the maximum shortest unweighted path between any two vertices.
Intuitively, this value of $t$ seems suitable since it allows for a transition probability between all nodes in the graph.
I.e., if $t \geq d$ then there is a probability of transitioning between every node pair in a random walk of length $t$.

We next demonstrate our pipeline on two simple examples to gain intuition before giving the results in Sec.~\ref{sec_results}.

\subsection{First Example: Comparing Distance Measurements} \label{ssec_example_one}

\begin{figure}
    \centering
    \includegraphics[width = \textwidth]{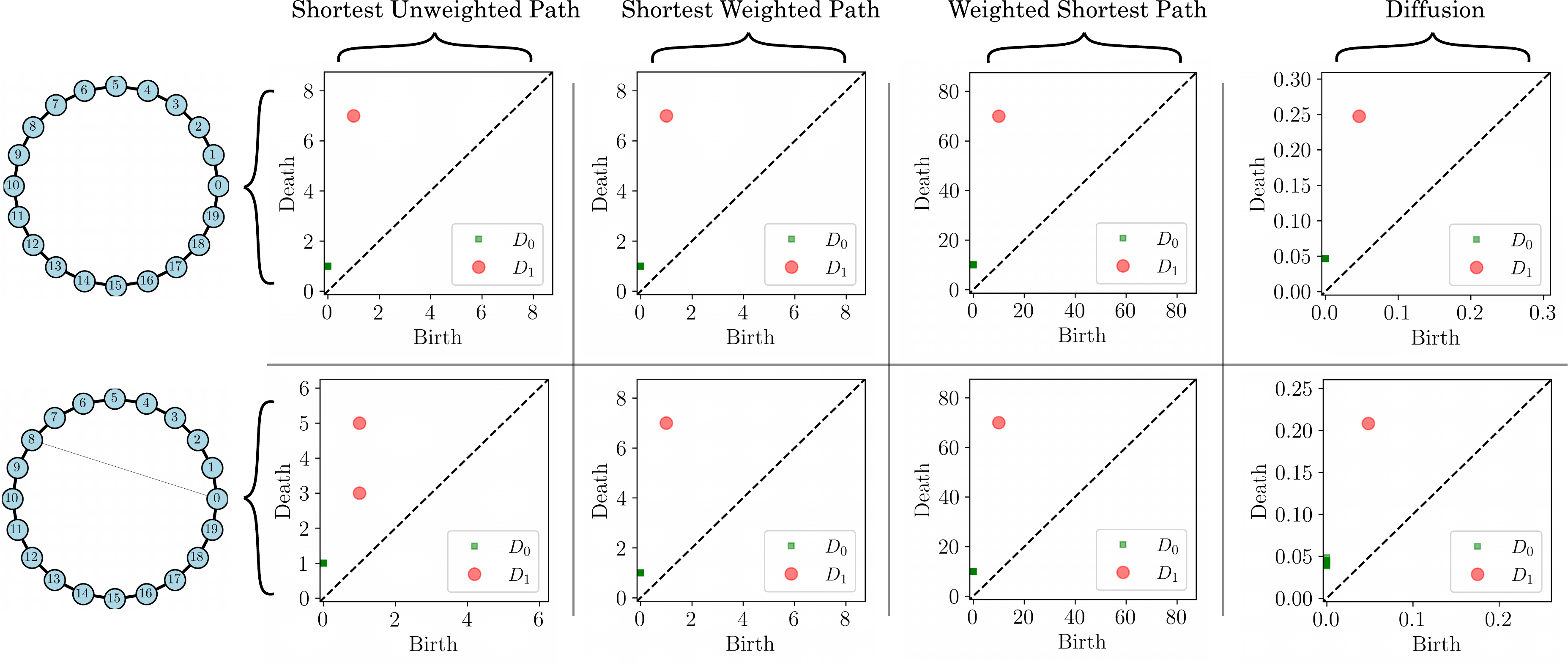}
    \caption{Two example weighted cycle graphs of weight 10 with the bottom row having an additional edge of weight one connecting nodes 0 and 8. The persistence diagram associated to each of the four distance methods are shown by column both both graphs.}
    \label{fig:example_cut_cycle_all_distances}
\end{figure}

To compare the original shortest unweighted path to the weight incorporating shortest path and diffusion distances, let us look at a simple example that highlights the issue previously mentioned with the unweighted shortest path not accounting for weight information.
In Fig.~\ref{fig:example_cut_cycle_all_distances} there are two graphs: on the top is a cycle graph with edge weights of 10 and on the bottom is the same cycle graph but with an additional single perturbation edge added between nodes 0 and 8 with a weight of 1.
This edge could be caused by additive noise, a perturbation to the underlying dynamical system, or simply a falsely added state transition in the OPN formation procedure.
If we implement the shortest unweighted path distance for calculating the persistent homology of the cycle graph we get a single significant point in the resulting persistence diagram as shown in the top left persistence diagram of Fig.~\ref{fig:example_cut_cycle_all_distances}.
However, adding the single, low-weighted edge splits the graph with the persistence diagram using the shortest unweighted path distance having two significant points in the persistence diagram (see bottom left diagram of Fig.~\ref{fig:example_cut_cycle_all_distances}).
This is due to the edge weight information being discarded when using the shortest path distance.

In comparison to the shortest unweighted path distance, the second, third, and fourth columns of Fig.~\ref{fig:example_cut_cycle_all_distances} show the persistence diagrams for both graphs using the shortest weighted path, weighted shortest path, and diffusion distances, respectively. For all three of these distance methods there is only a single one-dimensional point in the persistence diagrams for both graphs. Additionally, both the shortest weighted path and weighted shortest path have identical persistence diagrams for both graphs. This is due to the shortest weighted path between any two vertices never using the edge between vertices 0 and 8.
For the diffusion distance we also only have a single point in the persistence diagram for one-dimensional features.
This is caused by the weighted information being used in the diffusion distance calculation where the change in distance from the nodes 0 and 8 is not significantly changed from the addition of the perturbation edge connecting them since it has a low weight relative to the cycle and the transition probability distributions between vertices 0 and 8 are dissimilar. For calculating the diffusion distance in this example we used $t=2d$ walk steps.

\subsection{Second Example: Periodic and Chaotic Dynamics} \label{ssec_example_two}

The second example qualitatively demonstrates that persistence of the diffusion distance for OPNs can detect the dynamic state of a signal as either periodic or chaotic.
The example signal used here is from the Lorenz system defined as
\begin{equation}
\frac{dx}{dt}   = \sigma (y-x), \: \frac{dy}{dt}   = x (\rho -z) - y, \: \frac{dz}{dt}   = xy - \beta z.
 \label{eq:lorenz}
\end{equation}
The system was simulated with a sampling rate of 100 Hz and system parameters $\sigma = 10.0$, $\beta = 8.0 / 3.0$, and $\rho = 180.1$ for a periodic response or $\rho = 181.0$ for a chaotic response. This system was solved for 100 seconds with only the last 20 seconds used to avoid transients.

\begin{figure}
    \centering
    \includegraphics[width = 0.89\textwidth]{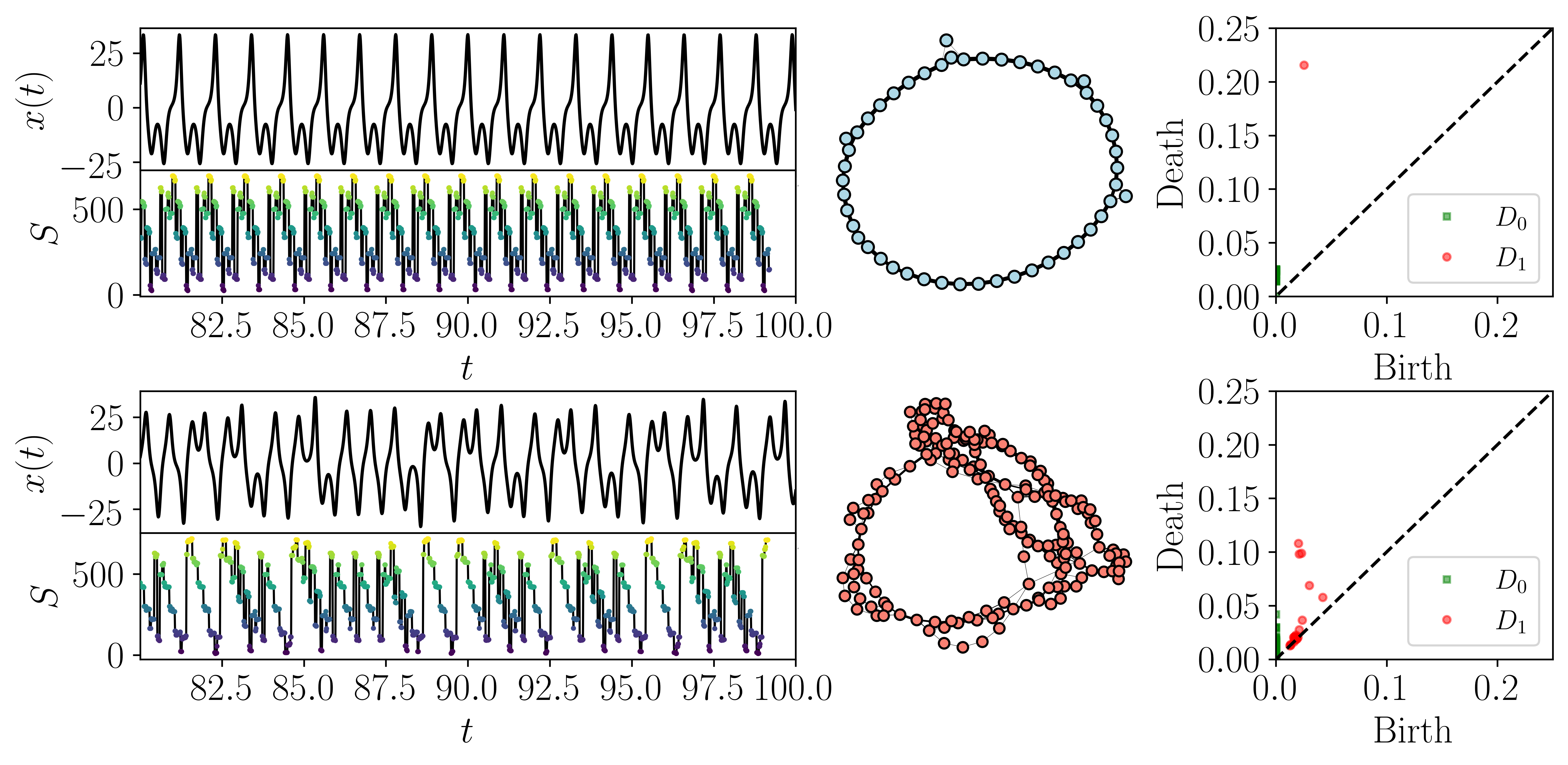}
    \caption{A comparison of the resulting persistence diagrams for an OPN formed from a periodic and chaotic signal from the Lorenz system. \revision{In each case, a function $x(t)$ is shown, and the ordinal partition states for a sliding window are shown below as $S(t)$.  The resulting networks are shown in the middle column and the resulting persistence diagrams are drawn at right. Note that the the ordinal parition states labeled $S$ do not have a natural ordering so an arbitrary one is chosen and shown.} }
    \label{fig:example2_periodic_vs_chaotic_lorenz}
\end{figure}

Figure~\ref{fig:example2_periodic_vs_chaotic_lorenz} shows the resulting Lorenz system simulation signals $x(t)$ for periodic (top row of figure) and chaotic (bottom row of figure) dynamics with the corresponding ordinal partition state sequence $S$ using dimension $n=6$ and $\tau = 17$ selected using multi-scale permutation entropy~\cite{Myers2020c}, OPN, and persistence diagram. This example result demonstrates that the persistence diagram for a periodic signals tend to have one or few significant points in the persistence diagram of one dimensional features $D_1$ representing the cyclic nature of the signal. On the other hand, the $D_1$ for chaotic signal has many significant points representing the entanglement of the OPN.

The other distance methods also demonstrate similar behavior when comparing the resulting persistence diagrams from periodic and chaotic dynamics.

\section{Results}  \label{sec_results}

In this section we discuss the empirical results on the dynamic state detection capabilities and stability of the persistent homology of ordinal partition networks using the distance methods for incorporating weight information.

\subsection{Dynamic State Detection}
\begin{figure}
    \centering
    \begin{minipage}[t]{0.42\textwidth}
        \centering
        \includegraphics[width=\linewidth]{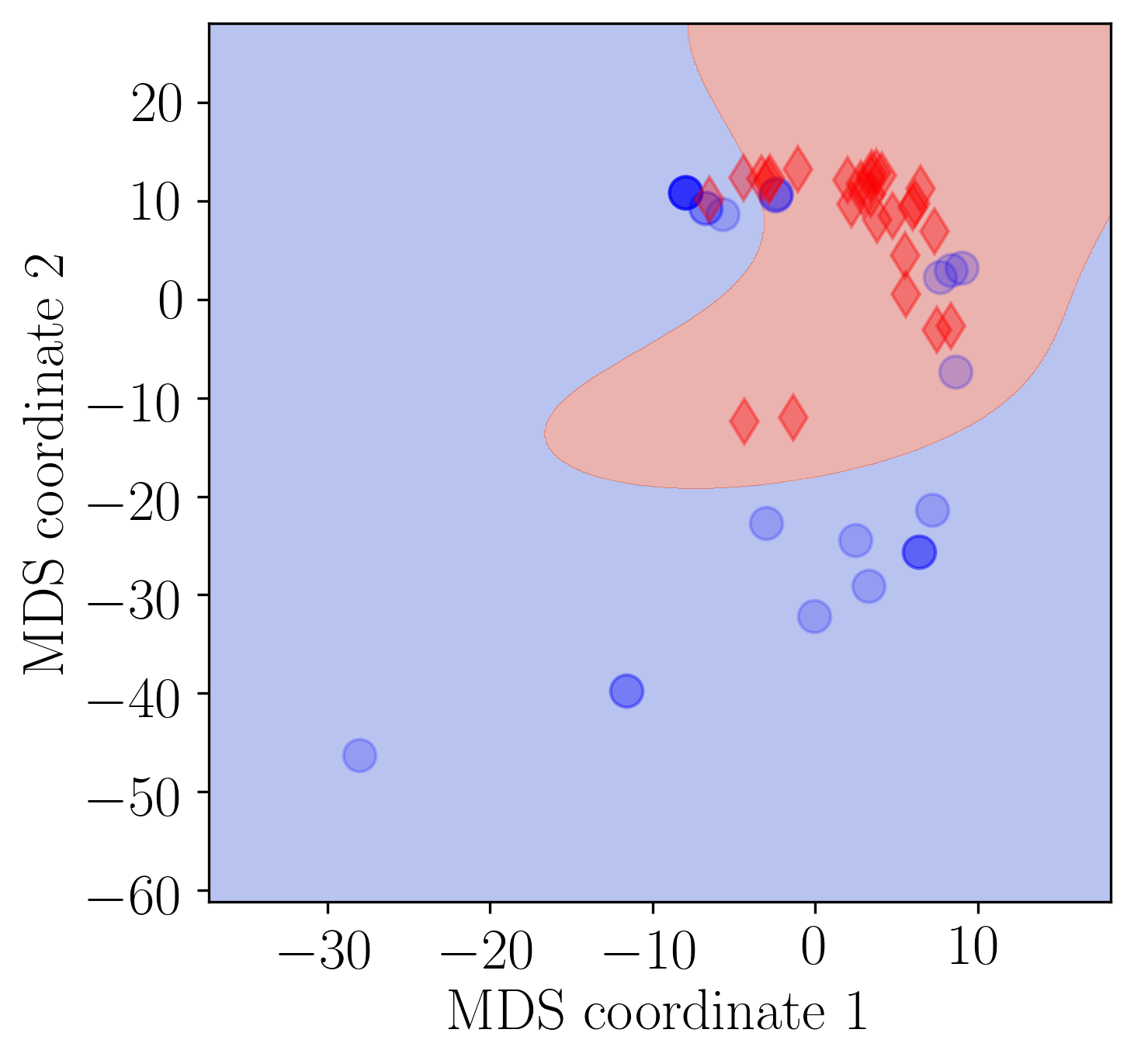}
         {Shortest unweighted path distance.}
    \end{minipage}
    \begin{minipage}[t]{0.42\textwidth}
        \centering
        \includegraphics[width=\linewidth]{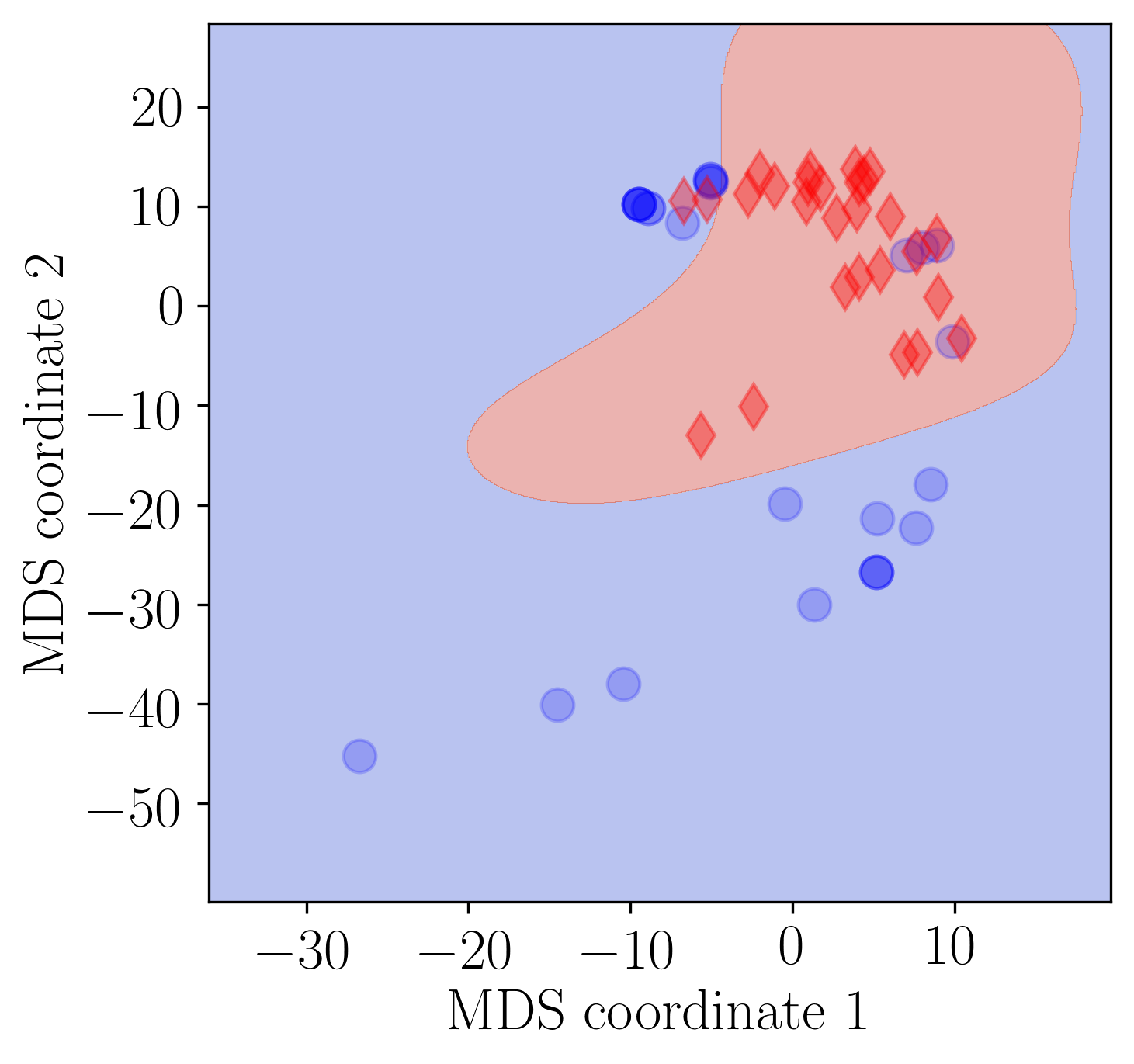}
         {Shortest weighted path distance.}
    \end{minipage}
    \begin{minipage}[t]{0.415\textwidth}
        \centering
        \includegraphics[width=\linewidth]{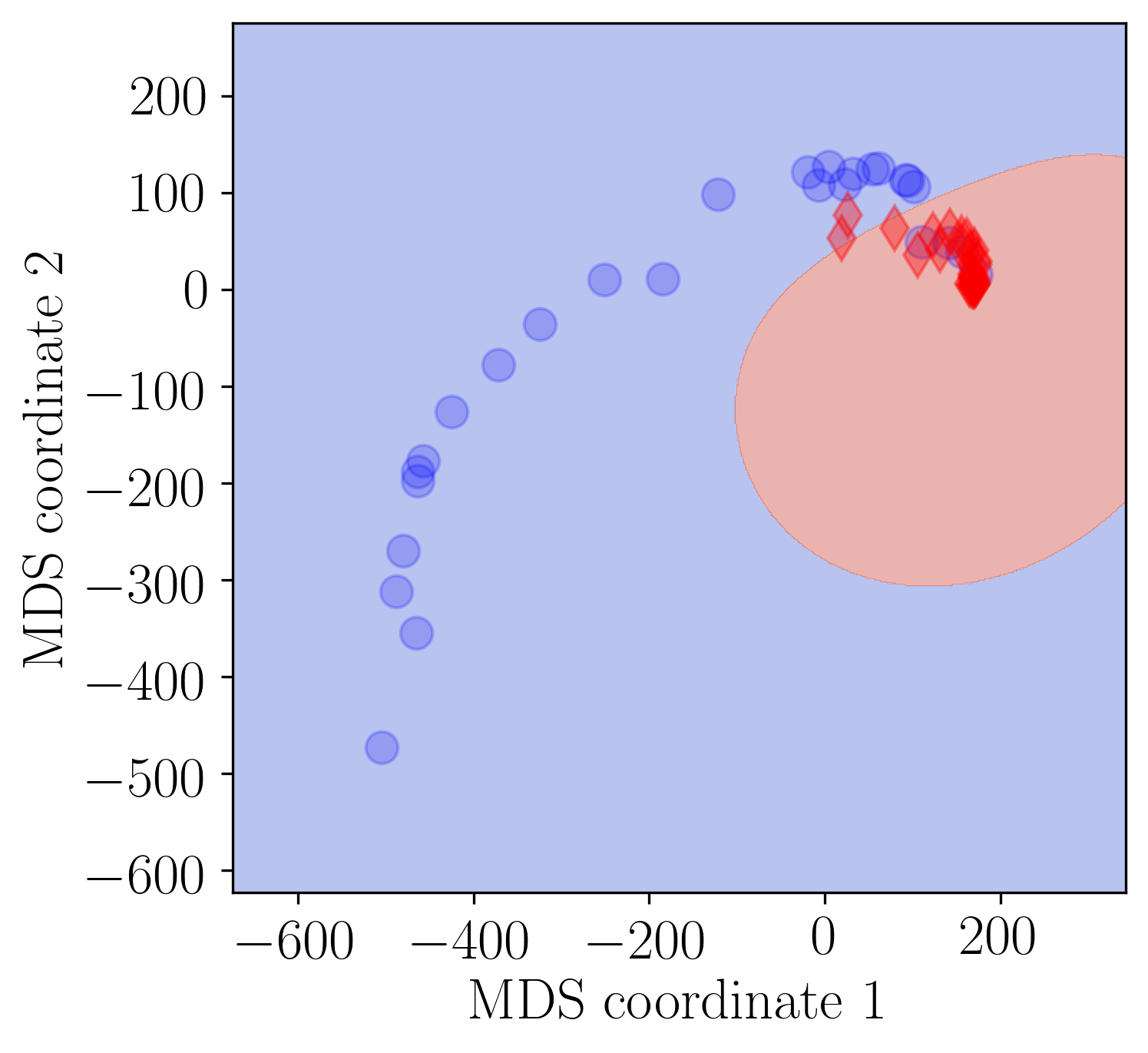}
         {Weighted shortest path distance}
    \end{minipage}
    \begin{minipage}[t]{0.42\textwidth}
        \centering
        \includegraphics[width=\linewidth]{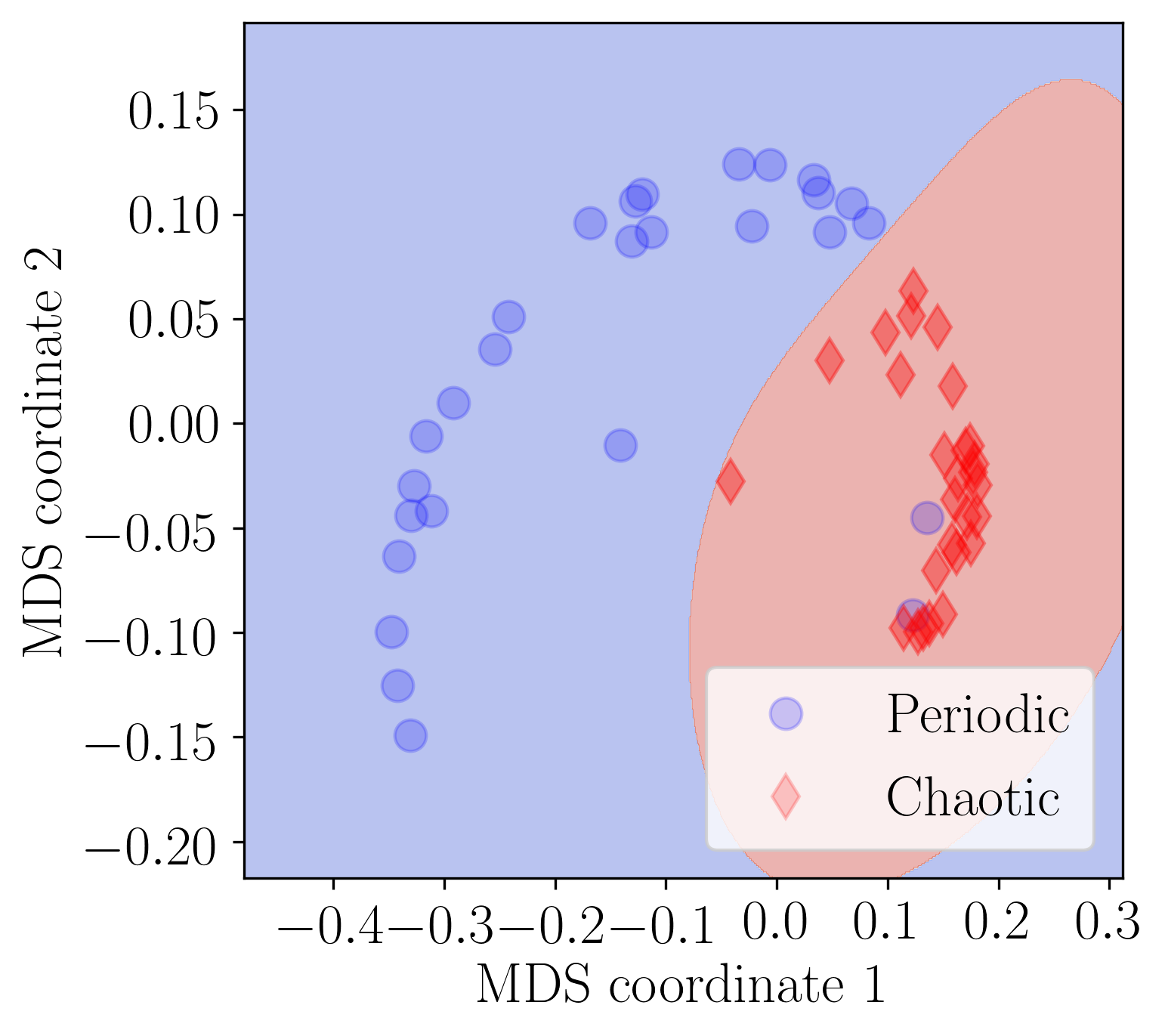}
         {Lazy diffusion distance}
    \end{minipage}
    \caption{Comparison between the (a) shortest unweighted path, (b) shortest weighted path, (c) weighted shortest path, and (d) lazy diffusion distances using a two dimensional MDS projection (random seed 42) of the bottleneck distances between persistence diagrams of the chaotic and periodic dynamics with an SVM radial bias function kernel separation. \revision{Note that the symbols are translucent so that overlapping points can be seen.}}
    \label{fig:MDS_all}
\end{figure}

To determine the viability of the persistence diagram for categorizing the dynamic state of a signal using the persistent homology of the shortest weighted path, weighted shortest path, and diffusion distances compared to the shortest unweighted path distance we use the lower dimensional projection of the persistence diagrams.
Specifically, we implemented the Multi-Dimensional Scaling (MDS) projection to two dimensions using the bottleneck distance matrix for our 23 systems (see Table~\ref{tab:systems} for a list). These systems were simulated from the dynamical systems module in the Python package \texttt{Teaspoon} with details on the simulations provided in Appendix~\ref{app:data}.
We then use a Support Vector Machine (SVM) with a Radial Basis Function (RBF) kernel to delineate periodic and chaotic dynamics based on the two dimensional MDS projection.
The SVM fit was done using default parameters for the \texttt{SKLearn} SVM package in Python.
The resulting separations shown in Fig.~\ref{fig:MDS_all} are for periodic and chaotic dynamics using the persistence diagrams with the following distances: shortest unweighted path (Fig.~\ref{fig:MDS_all}~a), shortest weighted path (Fig.~\ref{fig:MDS_all}~b), weighted shortest path (Fig.~\ref{fig:MDS_all}~c), and diffusion distance (Fig.~\ref{fig:MDS_all}~d).
\revision{
    We again reiterate (see the discussion of Sec.~\ref{ssec:MDS}) that these figures cannot be viewed as proper supervised learning testing as MDS does not allow for determining an embedding from an unseen test point (in this case, a persistence diagram). 
    Rather, these figures are useful for a visual proof of concept ensuring that the persistence diagrams representing different states are far apart in persistence diagram space. 
    Future work will be required in order to convert this setting into a point statistic to match to state type; however this work can be used in the case where a distribution of time series is in use and separation by clustering in persistence diagram space has potential for determining differences in the states.
}

The accuracy for each SVM kernel are provided as the standard distance percent accuracy in Table~\ref{tab:summarized_accurcies}.
\begin{table}[h!]
\centering
\caption{Accuracies of vaious distance methods (standard and normalized).}
\label{tab:summarized_accurcies}
\begin{tabular}{ccc}
\hline
\multirow{2}{*}{\textbf{Distance Method}} & \multicolumn{2}{c}{\textbf{Percent Accuracy  (\%)}} \\ \cline{2-3}
 & Standard Distance & Normalized Distance \\ \hline
Shortest unweighted path & 80.7 $\pm$ 1.5 & 91.9 $\pm$ 0.9 \\
Shortest weighted path & 88.9 $\pm$ 0.0 & 95.9 $\pm$ 0.8 \\
Weighted shortest path & 88.9 $\pm$ 0.0 & 92.6 $\pm$ 0.0 \\
Lazy diffusion distance & 95.0 $\pm$ 0.9 & 91.0 $\pm$ 0.7 \\
\hline
\end{tabular}
\end{table}
Based on this initial analysis it is clear that the diffusion distance significantly outperforms the other distance methods with an accuracy of 95.0\% $\pm$ 0.9\%. We theorize that one reason for the increased performance when using the diffusion distance is in how it tends to normalize the scale of the persistence diagram. Specifically, when comparing the 23 dynamical systems, the maximum lifetimes for $t = 2d$ walk steps ranges from 0.08 to 0.21 with a mean of 0.147 and standard deviation of 0.042 or 28.6\% of the average. In comparison, the maximum lifetimes for the shortest unweighted path distance range from 2 to 24 with an average of 9.38 and standard deviation of 6.36 or 67.8\% of the average. This demonstrates that the persistence diagrams from the diffusion distance calculation tends to be more consistent in magnitude.
We can further show this relationship using the cycle graph $G_{\rm cycle}(n)$, where $n$ as the number of nodes is increased from 2 to 500 with the maximum persistence calculated for each graph (see Appendix Section~\ref{app:cycle_graph}). In comparison to the shortest path distances, this result shows that the persistence of the cycle graph does not continue to grow with a larger cycle graph when using the diffusion distance and trends to a plateau.

\begin{figure}[h!]
    \centering
    \begin{minipage}[t]{0.42\textwidth}
        \centering
        \includegraphics[width=\linewidth]{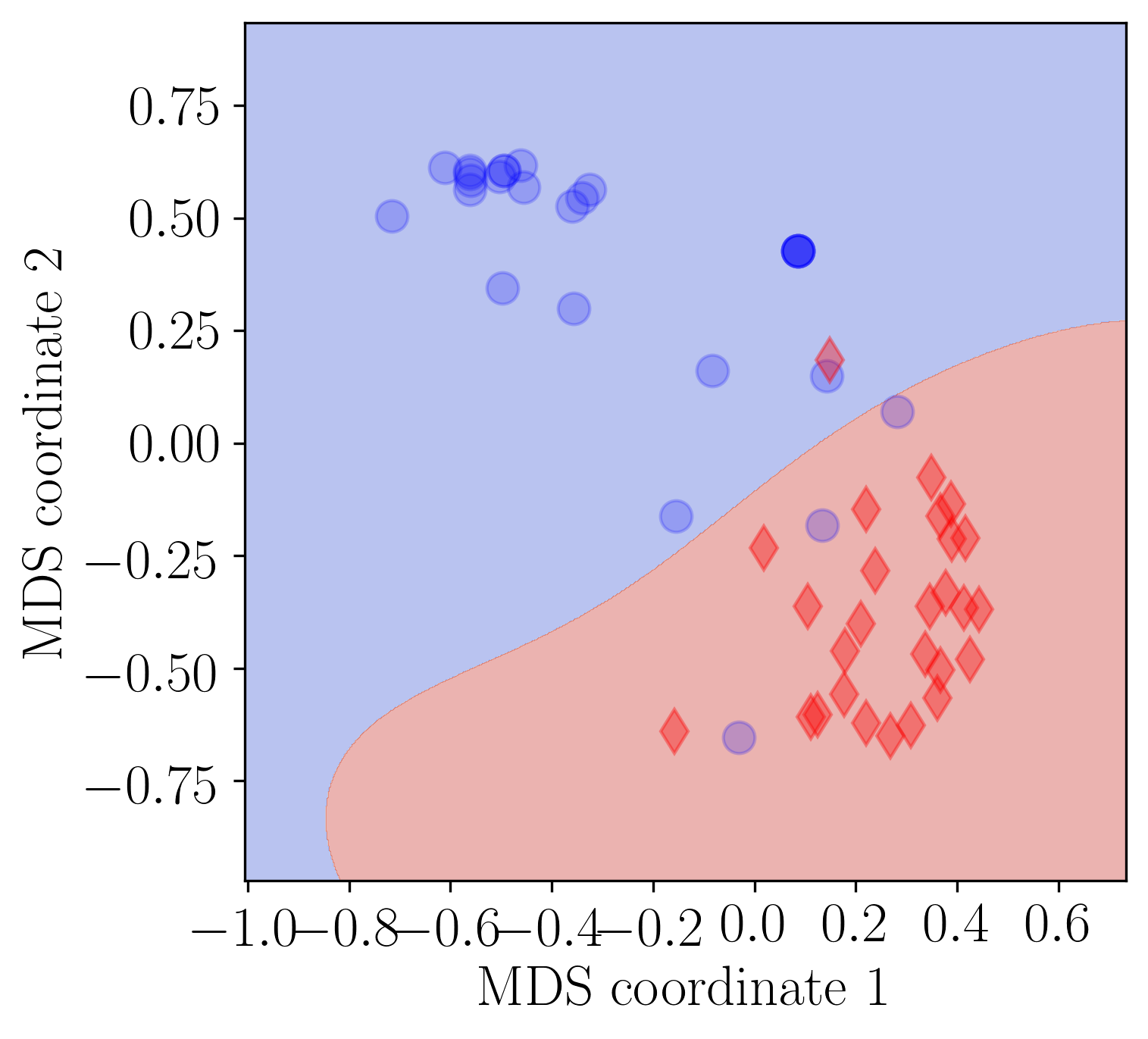}
         {Normalized shortest unweighted path distance.}
    \end{minipage}
    \begin{minipage}[t]{0.42\textwidth}
        \centering
        \includegraphics[width=\linewidth]{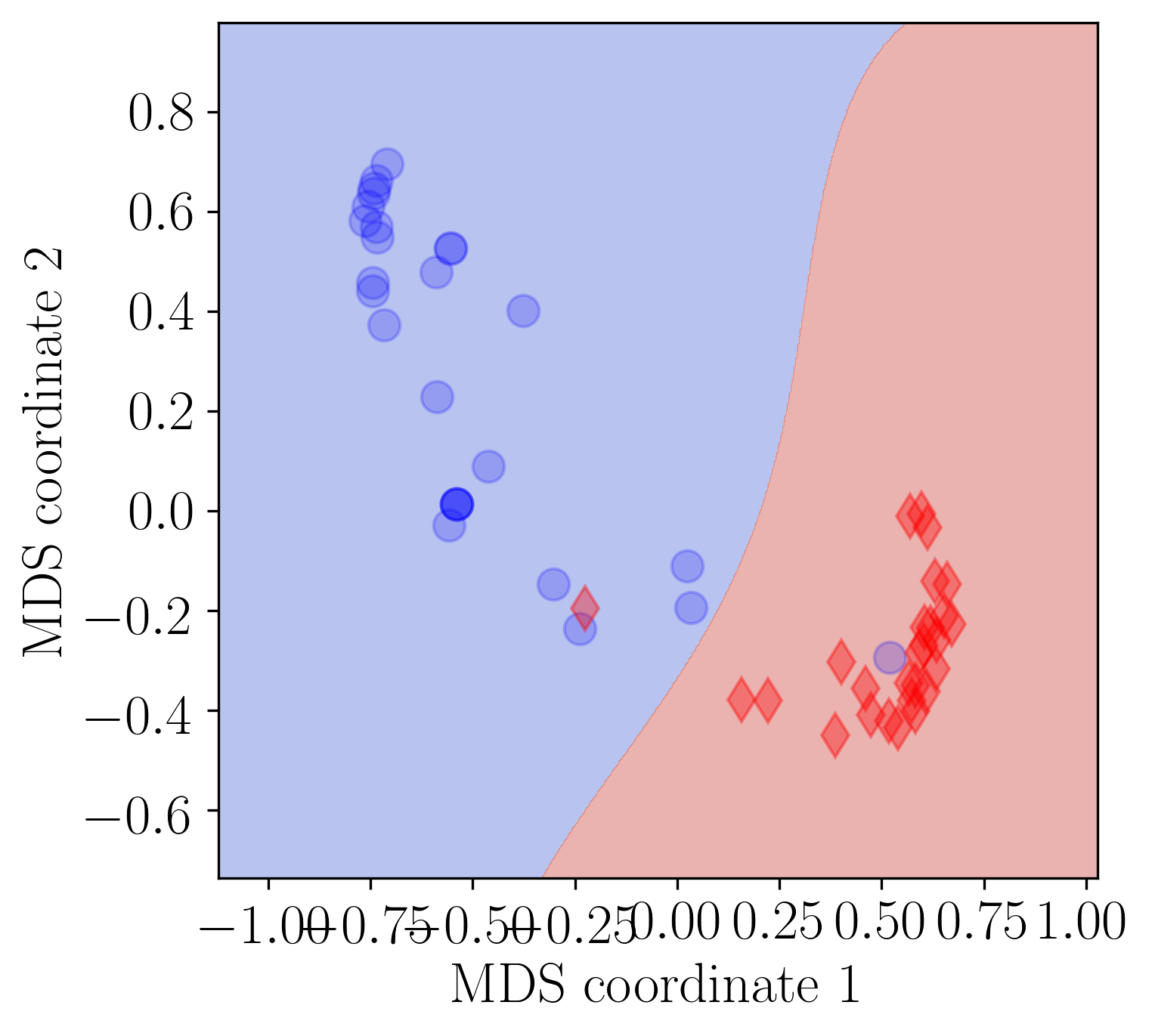}
         {Normalized shortest weighted path distance.}
    \end{minipage}
    \begin{minipage}[t]{0.415\textwidth}
        \centering
        \includegraphics[width=\linewidth]{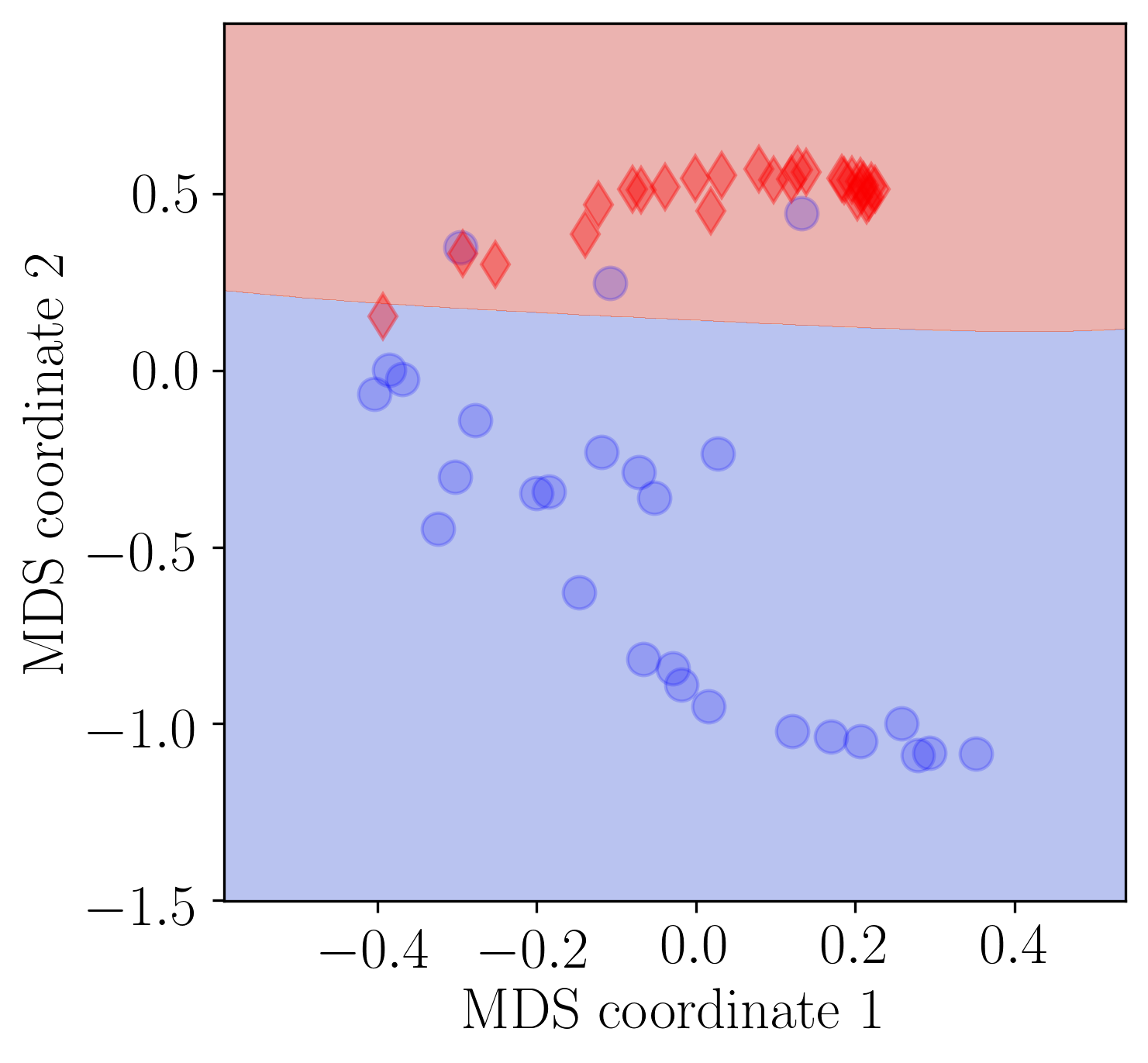}
         {Normalized weighted shortest path distance}
    \end{minipage}
    \begin{minipage}[t]{0.42\textwidth}
        \centering
        \includegraphics[width=\linewidth]{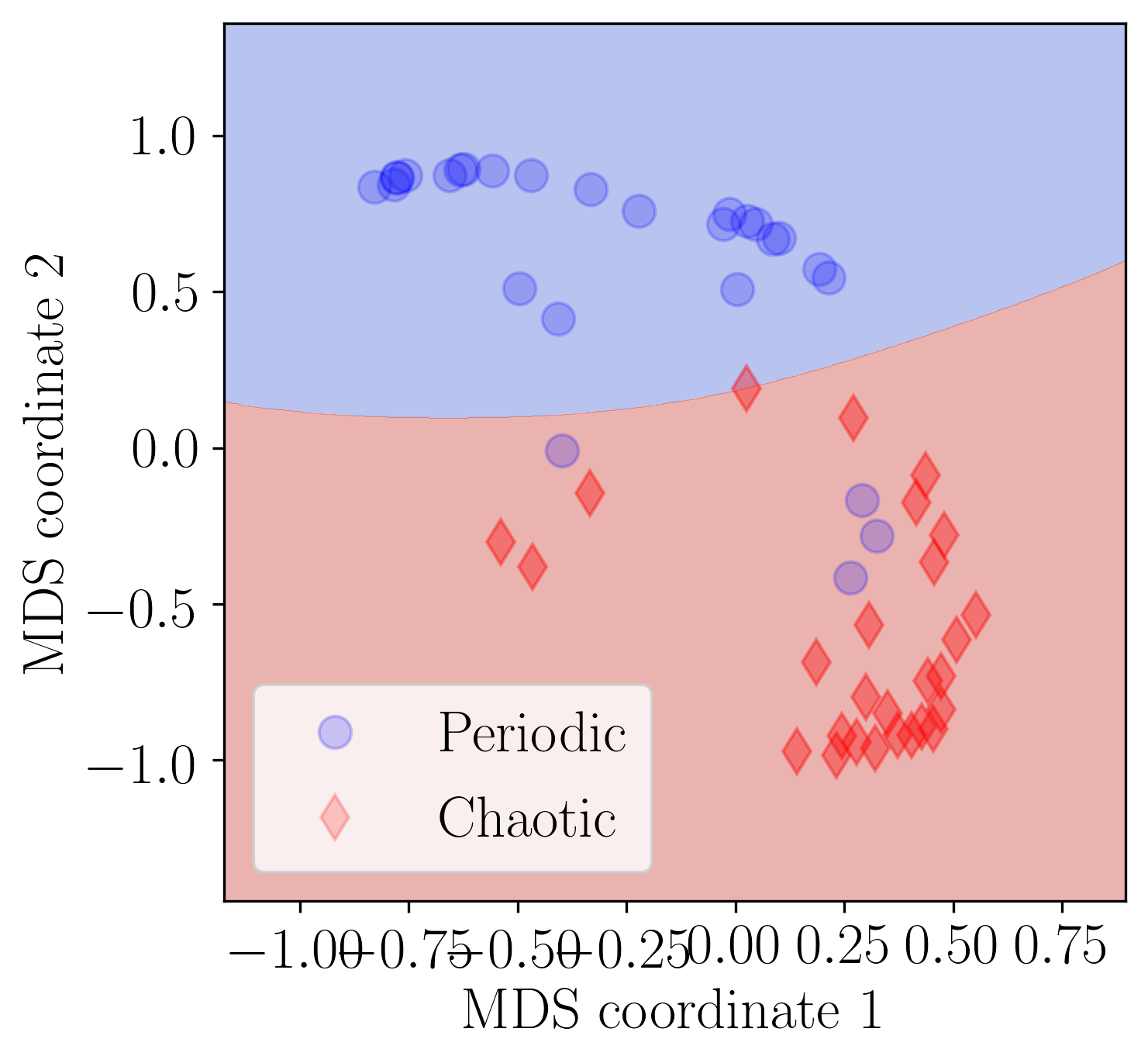}
         {Normalized lazy diffusion distance}
    \end{minipage}
    \caption{Two dimensional MDS projection (random seed 42) of the bottleneck distances between persistence diagrams of the chaotic and periodic dynamics with an SVM radial bias function kernel separation. Distance matrices are normalized for the (a) shortest unweighted path, (b) shortest weighted path, (c) weighted shortest path, and (d) lazy diffusion distances. \revision{Note that the symbols are translucent so that overlapping points can be seen.}}
    \label{fig:MDS_all_normalized}
\end{figure}

The accuracy improvements from the natural normalization characteristics of the diffusion distance can be extended to the other distances by normalizing their respective distance matrix as
\begin{equation}
D^{*} = \frac{D}{\max_{i, j}D(i, j)},
\end{equation}
where $D^*$ is the normalized distance matrix where $\max_{i, j}D^*(i, j) = 1$. As theorized, repeating the MDS and SVM analysis done in Fig.~\ref{fig:MDS_all} with the normalized distances (see Fig.~\ref{fig:MDS_all_normalized}) improves the accuracy of the other distances.

Visually, the SVM kernel separates the dynamic states more accurately when using the normalized distances. When this analysis was repeated for 100 random seeds we found that there was an improvement in accuracy for each of the shortest path distances, but a slight decrease in accuracy when using the diffusion distance (see Table~\ref{tab:summarized_accurcies}. Specifically, we found approximately a 3\% decrease in accuracy when using the diffusion distance, but a significant improvement in all the shortest path distances. Overall the best performance in terms of accuracy is found when using the normalized shortest weighted path with a $95.9\% \pm 0.8\%$ accuracy. This shows the importance of normalizing the distance matrix when using SVM kernels for dynamic state detection using an MDS projection.

\subsection{Stability Analysis}
One drawback to using MDS in our setting is that it cannot be used for true supervised learning as data points not in the original training set cannot be assigned a projection after the fact.
We can at least analyze how sensitive the bottleneck distance between persistence diagrams is to differences in the input time series, showing that the results are resilient to noise.
While we would like to be able to provide a stability proof in the spirit of \cite{Cohen-Steiner2006}, such an investigation is outside the scope of this work (see further discussion in Sec.~\ref{sec_conclusion}).

Instead we use an empirical study of the stability of the bottleneck distance using the same 23 systems with the periodic signals (both dissipative autonomous and driven).
Specifically, we tested the stability by adding bounded Gaussian noise to the signal.
The noise had Signal to Noise Ratios (SNR) from $\infty$ (no noise) to 15 dB (extremely noisy).
The additive noise followed a zero-mean Gaussian distribution that was truncated at three standard deviations from the mean and set $\e = 6 \sigma$. To make a fair comparison between each of the distance methods in terms of stability and sensitivity to noise we normalize the bottleneck distance as
\begin{equation}
d^*_B(D_1, D_1^{\e}) = \frac{d_B(D_1, D_1^{\e})}{\frac{1}{2}\sum_{x \in D_1} {\rm pers}(x)},
\end{equation}
where $d_B$ is the bottleneck distance function and $D_1$ and $D_1^{\e}$ are the noise free and noise contaminated one-dimensional persistence diagrams, respectively.

\begin{figure}[h!]
    \centering
    \includegraphics[width=0.67\textwidth]{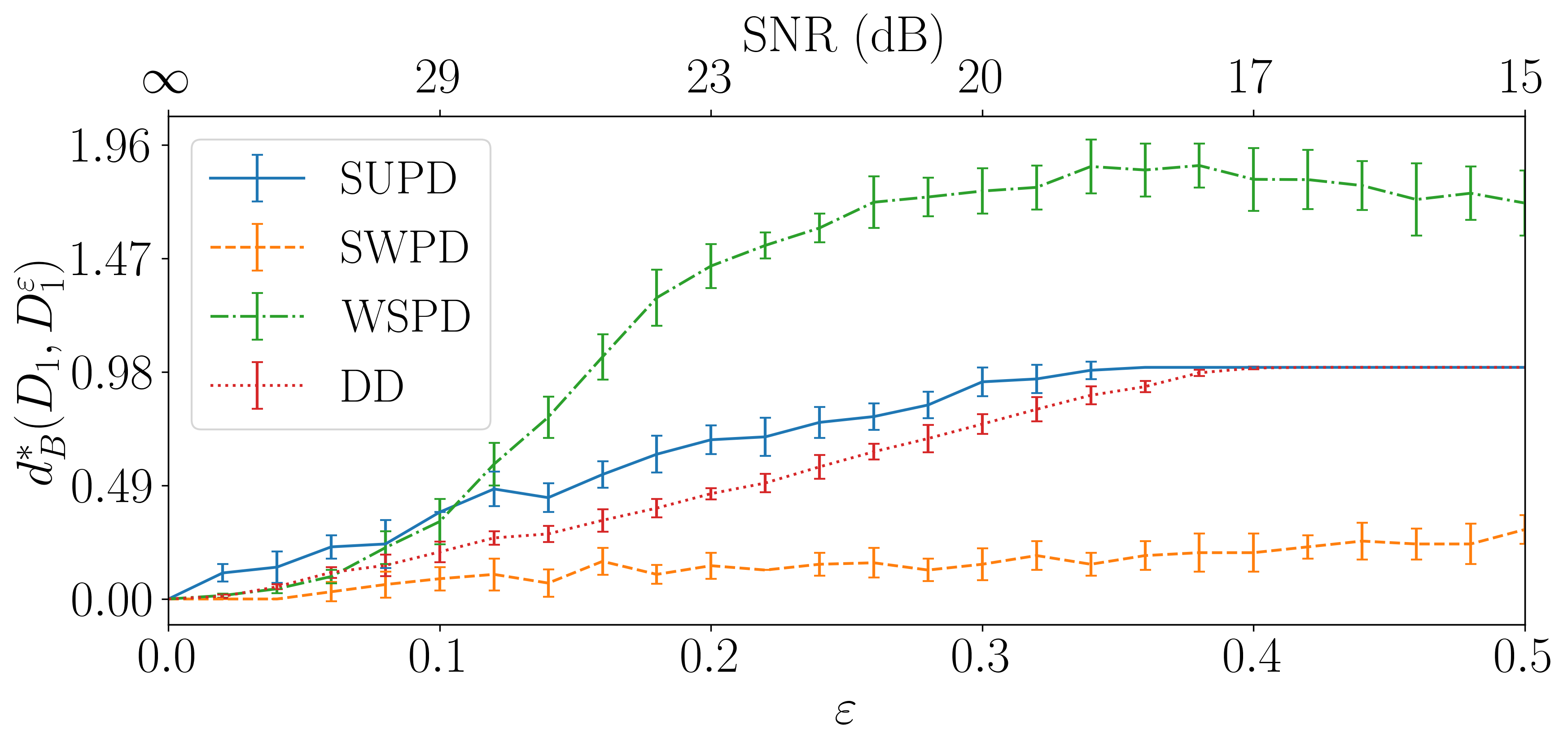}
    \caption{Bottleneck distance stability analysis of the periodic Lorenz system (see Eq.~\eqref{eq:lorenz}) with standard deviation normalized signal and bounded ($\varepsilon = 6\sigma$) Gaussian additive noise. Analysis shows stability results using Shortest Unweighted Path Distance (SUPD), Shortest Weighted Path Distance (SWPD), Weighted Shortest Path Distance (WSPD), and Diffusion Distance (DD).}
    \label{fig:stability_example}
\end{figure}

Figure~\ref{fig:stability_example} provides a demonstrative example of the effects of noise and the stability of the persistence diagram for the Lorenz system.
The persistence diagrams as $\e$ is increased are drawn overlaid in Fig.\ref{fig:stability_example}~(b)
In Fig.\ref{fig:stability_example}, we see the bottleneck distance from the the noise free diagram to the noise contaminated diagram as the noise amplitude $\e$ is increased.
In the case of Lorenz, all four distance methods are stable with an approximately linear change in the bottleneck distance with respect to the noise level $\e$ for small levels of noise (less than 25 dB). Additionally, $d_B^*$ tends to plateau at noise levels greater than approximately 18 dB. This is due to the minimum pairing between diagrams matching to the diagonal. It is also clear the shortest weighted path distance is significantly less sensitive to additive noise with only slight changes in its normalized bottleneck distance as $\e$ is increased.

Some of these characteristics seen in the Lorenz systems seem to be consistent across all of the other 22 systems; see Appendix Section~\ref{appx:sec:stability} for similar figures for the remaining systems. The shortest weighted path distance tends to be the least sensitive to additive noise. Additionally, the bottleneck distance tends to plateau at approximately 20 dB for most systems. Most importantly, all of the distance methods tend to have an approximately linear relationship between $d_B^*$ and $\e$ for low levels of noise (SNR $\leq 25$ dB).
These results empirically demonstrate that the persistence diagram is stable in this setting for limited levels of additive noise.

Some characteristics that tend to be highly dependent on the system is the sensitivity of the shortest unweighted path, weighted shortest path, and diffusion distances to additive noise. For some systems (e.g. the Rabinocih Frabrikant attractor), the weighted shortest path distance is the least sensitive to high levels of additive noise, while in other systems (e.g. the Thomas cyclically symmetric attractor) the weighted shortest path distance is the most sensitive to additive noise. In most systems the diffusion distance and shortest unweighted path are comparably sensitive to additive noise.

\section{Conclusions}  \label{sec_conclusion}

In this work we investigated the viability of encoding the behavior of time series through the persistent homology of the weighted ordinal partition network.
Our results show that there is a significant improvement when using the weighted distance methods such as the shortest weighted path, weighted shortest path, and diffusion distance. 
These weighted distance methods incorporate information about the edge weights that gets lost when using the unweighted shortest path distance.
For our analysis we used the MDS projection and an SVM with an RBF kernel to separate periodic from chaotic dynamics with 23 continuous systems and the bottleneck distance matrix between each systems resulting persistence diagram.
By using a large set of dynamic systems we are able to empirically evaluate the performance for dynamic state detection.
\revision{We found an increase in dynamic state separation when using the diffusion distance over the shortest path distance.}
We also investigated the performance when using the normalized distances and found that this caused a significant improvement in the dynamic state detection accuracy for all of the shortest path distances.

Due to limitations of the MDS projection, we further studied the sensitivity to noise by investigating how the bottleneck distance to the noise-free diagram changes as we increase the noise added to the original time series.
While this work provides an interesting empirical study suggesting stability for the distance methods, an interesting question to consider is that of proving a stability theorem.
The general stability theorem for persistent homology~\cite{Cohen-Steiner2006} requires
\begin{equation}
	d_B({\rm diag}(f), {\rm diag}(g)) \leq  || f - g ||_{\infty},
\end{equation}
where $f$ and $g$ are functions on a topological space and $d_B$ is the bottleneck distance of the sublevel-set persistence diagram.
In our work we would attempt to apply this to our distance functions by showing
\begin{equation}
	d_B({\rm diag}(\mathbf{D}_t), {\rm diag}(\mathbf{D}^{'}_t)) \stackrel{?}{\le} \max_{i, j} | \mathbf{D}_t(i,j) - \mathbf{D}^{'}_t(i,j) |,
\end{equation}
where $\mathbf{D}^{'}$ is the noise free diffusion distance matrix.
However, due to the complexities of considering the effects of the choice of $t$ and the structure of the graph being based on the ordinal partition segmentation of the state space, the difficulty of proving the stability is outside of the scope of this work but would certainly be of interest.

Another direction for future work relates to modifications of the choice of network and how much of the network information is utilized.
In future work, we will investigate the directed version of the OPN and apply newly developed methods for incorporating directed edges into persistent homology.
We also hope to extend this work to other choices of networks for representing dynamical systems. 

\bibliographystyle{siamplain}
\bibliography{references}

\newpage
\appendix

\section{Additional Diffusion Distance Analysis}

\subsection{Persistence of Cycle Graph} \label{app:cycle_graph}
\begin{figure}[h]
    \centering
    \includegraphics[width = 0.54\textwidth]{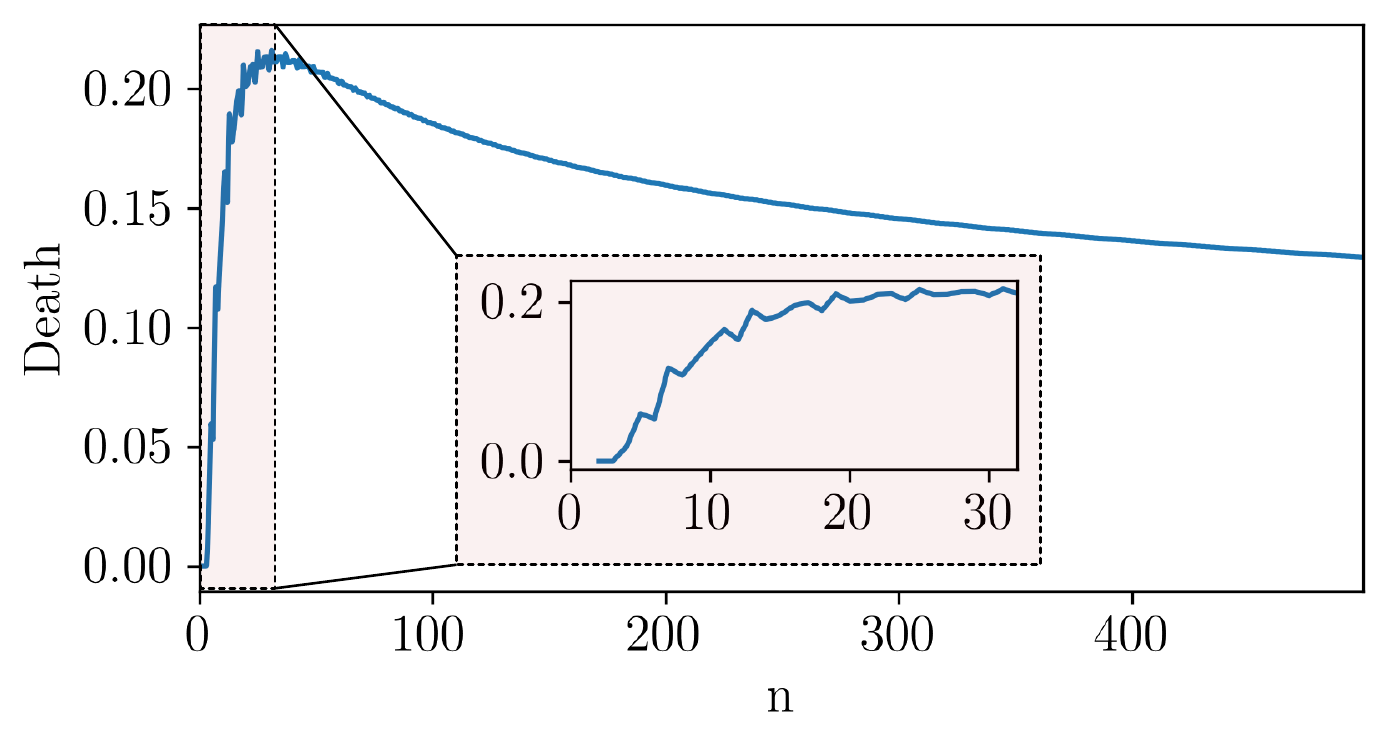}
    \caption{Numerical analysis of the maximum persistence of the cycle graph  $G_{\rm cycle}(n)$ with size $n$ when using diffusion distance with $t=2d$.}
    \label{fig:cycle_graph_size_analysis}
\end{figure}

The cycle graph on $n$ vertices is the graph $G=(V,E)$ with $V = \{v_1,\cdots,v_n\}$, and $E = \{v_iv_{i+1} \mid 1 \leq i <n \} \cup \{v_nv_1\}$; i.e.~it forms a closed path (cycle) where no repetitions occur except for the starting and ending vertices.
If we increase the number of nodes from 2 to 500 and calculate the maximum persistence or maximum lifetime, we find that it quickly reaches a maximum of $L_1 = 0.216$ at $n=32$, and then steadily declines seeming to approach a plateau as shown in Fig.~\ref{fig:cycle_graph_size_analysis}.
This is in comparison to the unweighted shortest path distance of the cycle graph which has a maximum persistence of $\lceil{n/3}\rceil - 1$ as shown in~\cite{Myers2019}.

\subsection{Analysis on Random Walk Steps} \label{appendix_section_on_t}

\begin{figure}[h]
    \centering
        \centering
        \includegraphics[width=\linewidth]{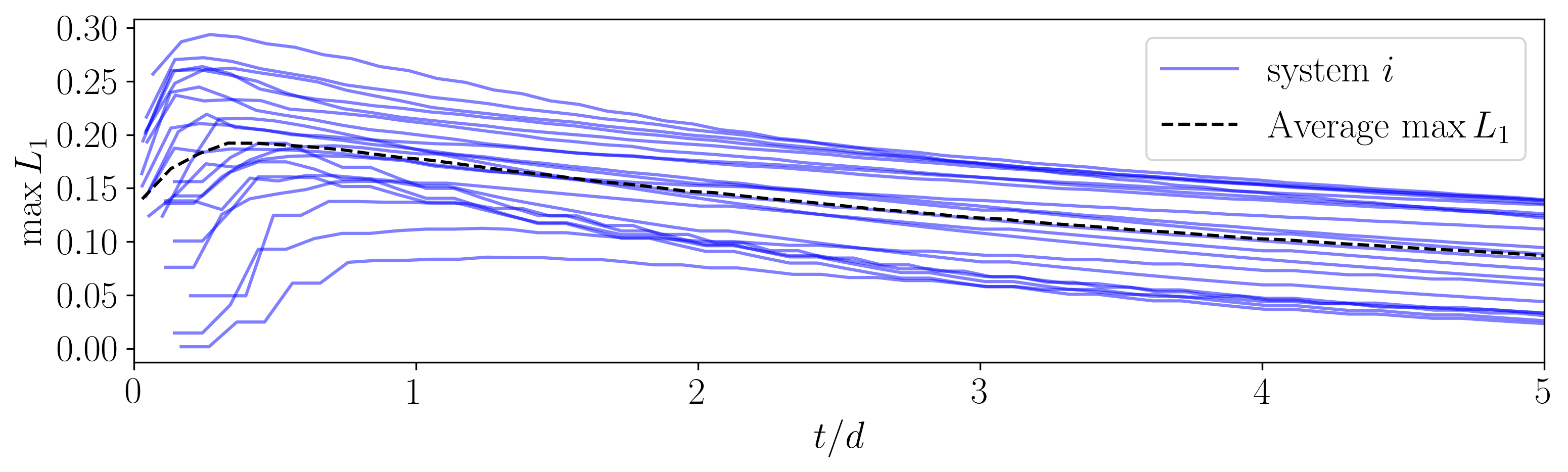}

    \vspace{0.01cm}
        \centering
        \includegraphics[width=\linewidth]{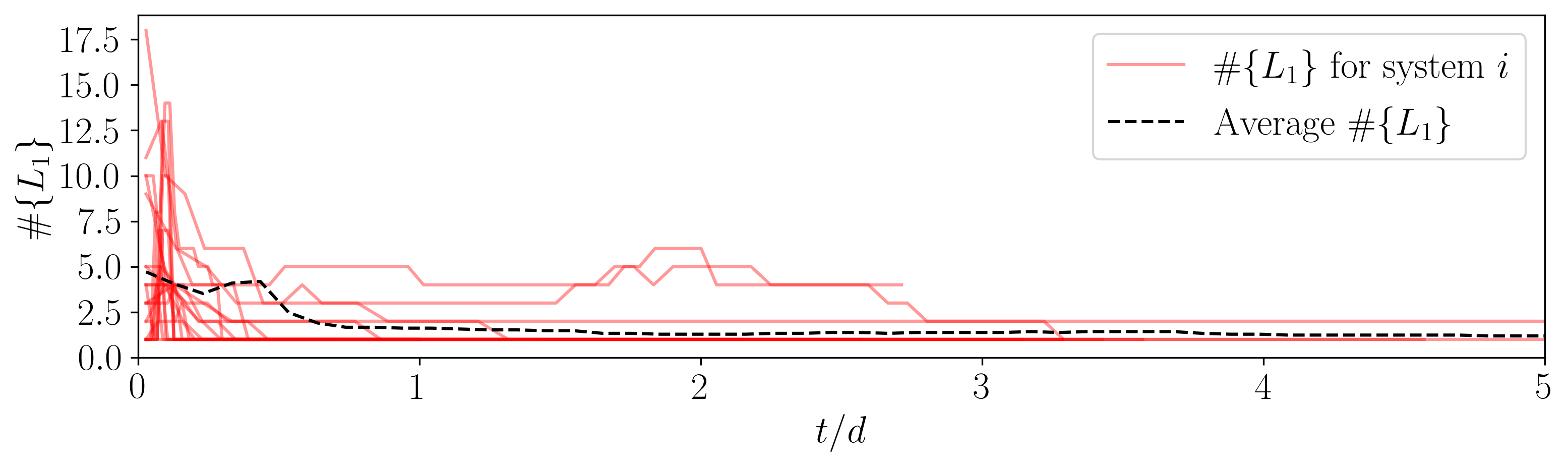}
    \caption{Comparison of $\max L_1$ and $\# \{ L_1 \}$ for each system and mean when varying $t$ in $P^t$ with respect to the diameter ($t \in [d, 5d]$).}
    \label{fig:t_analysis}
\end{figure}

In this section we vary the number of random walk steps $t$ with respect to the graph diameter $d$ to determine how many steps is suitable for calculating the persistent homology based on the diffusion distance. We vary $t/d$ from $1$ to $5$ as shown in Fig.~\ref{fig:t_analysis}. To decide on the optimal $t$ we calculate the maximum lifetime and number of persistence pairs in each resulting persistence diagram for each of the 23 dynamical systems investigated in this work. Additionally, the average for both the maximum lifetime and number of lifetimes is plotted as shown in Fig.~\ref{fig:t_analysis}.

Based on the each systems maximum lifetimes, a suitable value for $t$ should be greater than $d$ based on having a $t$ large enough that each system reaches a maximum of the $\max (L_1)$. We can also note that the number of persistence pairs or lifetimes in the persistence diagram does not stabalize for the majority of systems until approximately $t = 2d/3$. This again supports a minimum suggest $t > d$. The only downfall of larger values of $t$ is that the maximum lifetime tends to diminish as shown in the $\max (L_1)$ figure. Therefor, we conclude that a suitable $t$ should be within the range $d < t < 3d$. In this work we chose $t = 2d$.

\section{Data} \label{app:data}

In this work we heavily rely on a 23 dynamical systems commonly used in dynamical systems analysis. All of these systems are continuous flow opposed to maps. The 23 systems are listed in Table~\ref{tab:systems}. The equations of motion for each systems can be found in the python topological signal processing package \texttt{Teaspoon} under the module \textit{MakeData} \url{https://lizliz.github.io/teaspoon/}. Specifically, these systems are described in the dynamical systems function of the make data module~\cite{Myers2020}.

Each system was solved to have a time delay $\tau = 50$, which was estimated from the multiscale permutation entropy method~\cite{Myers2020c}. The signals were simulated for $750 \tau /f_s$ seconds with only the last fifth of the signal used to avoid transients. It should be noted that we did not need to normalize the amplitude of the signal since the ordinal partition network is not dependent on the signal amplitude.

\newpage
\section{Additional Figures for Stability Analysis}
\label{appx:sec:stability}
The following figures are additional empirical analysis of the stability of the persistent homology of weighted ordinal partition networks. The list of systems is shown in Table~\ref{tab:systems}.

\begin{table}[h!]
\centering
\caption{Continuous dynamical systems used in this work.}
\label{tab:systems}
\begin{tabular}{ll}
\textbf{Autonomous Flows} & \textbf{Driven Dissiptive Flows} \\ \hline
Lorenz & Driven Van der Pol Oscillator \\
Rossler & Shaw Van der Pol Oscillator \\
Double Pendulum & Forced Brusselator \\
Diffusionless Lorenz Attractor & Ueda Oscillator \\
Complex Butterfly & Duffing Van der Pol Oscillator \\
Chen's System & Base Excited Magnetic Pendulum \\
ACT Attractor &  \\
Rabinovich Frabrikant Attractor &  \\
Linear Feedback Rigid Body Motion System &  \\
Moore Spiegel Oscillator &  \\
Thomas Cyclically Symmetric Attractor &  \\
Halvorsen's Cyclically Symmetric Attractor &  \\
Burke Shaw Attractor &  \\
Rucklidge Attractor &  \\
WINDMI &  \\
Simplest Cubic Chaotic Flow & \\
\hline
\end{tabular}
\end{table}

\begin{figure}[h]
    \centering
    \begin{minipage}[t]{\myfiguresizeB\textwidth}
        \centering
        \includegraphics[width=\linewidth]{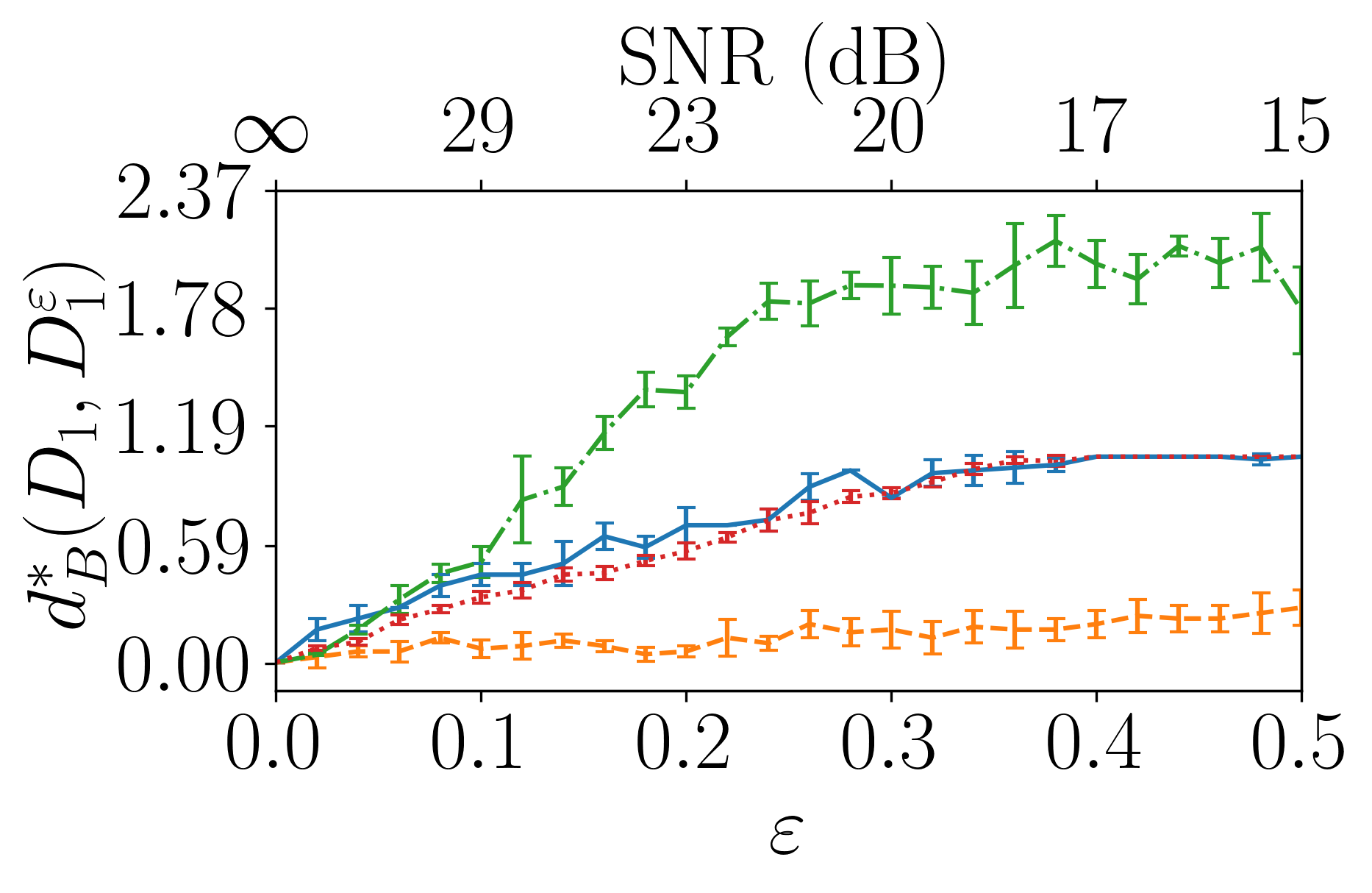}
         {Lorenz Attractor } %
    \end{minipage}
	\hfill
	\begin{minipage}[t]{\myfiguresizeB\textwidth}
        \centering
        \includegraphics[width=\linewidth]{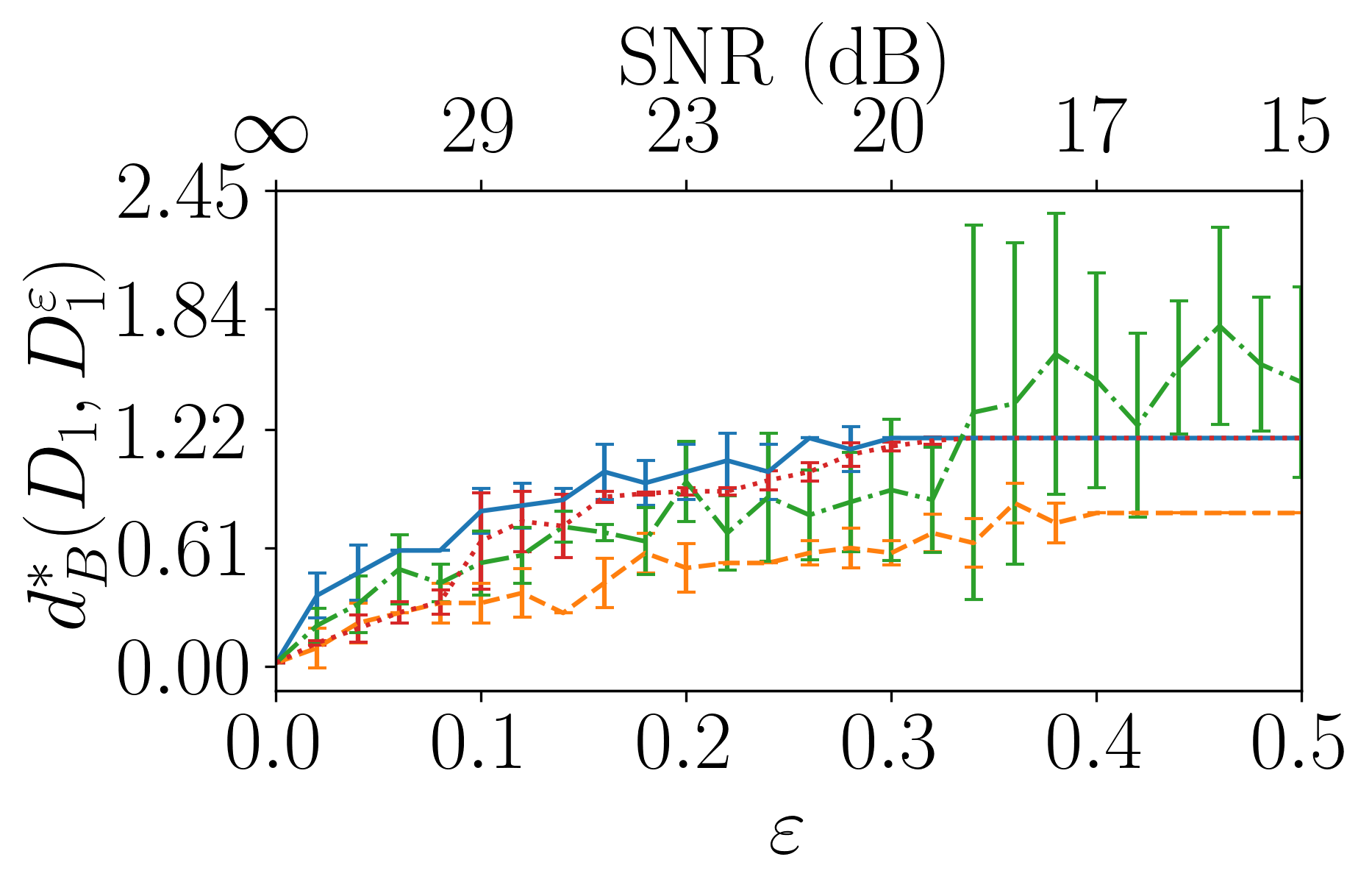}
         {Rossler Attractor } %
    \end{minipage}
    \hfill
    \begin{minipage}[t]{\myfiguresizeB\textwidth}
        \centering
        \includegraphics[width=\linewidth]{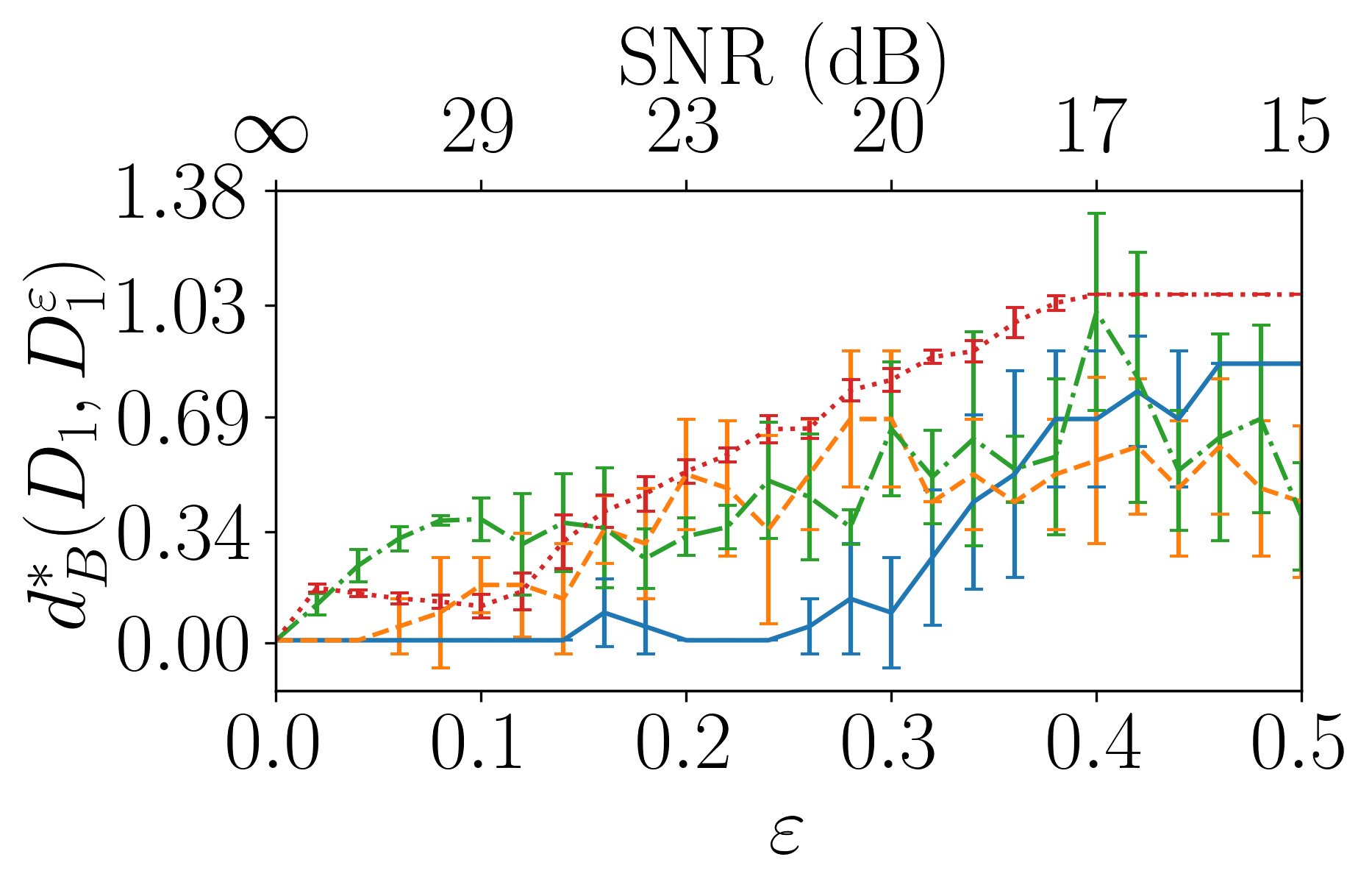}
         {Double Pendulum } %
    \end{minipage}

    	\vspace{0.2cm}

    \begin{minipage}[t]{\myfiguresizeB\textwidth}
        \centering
        \includegraphics[width=\linewidth]{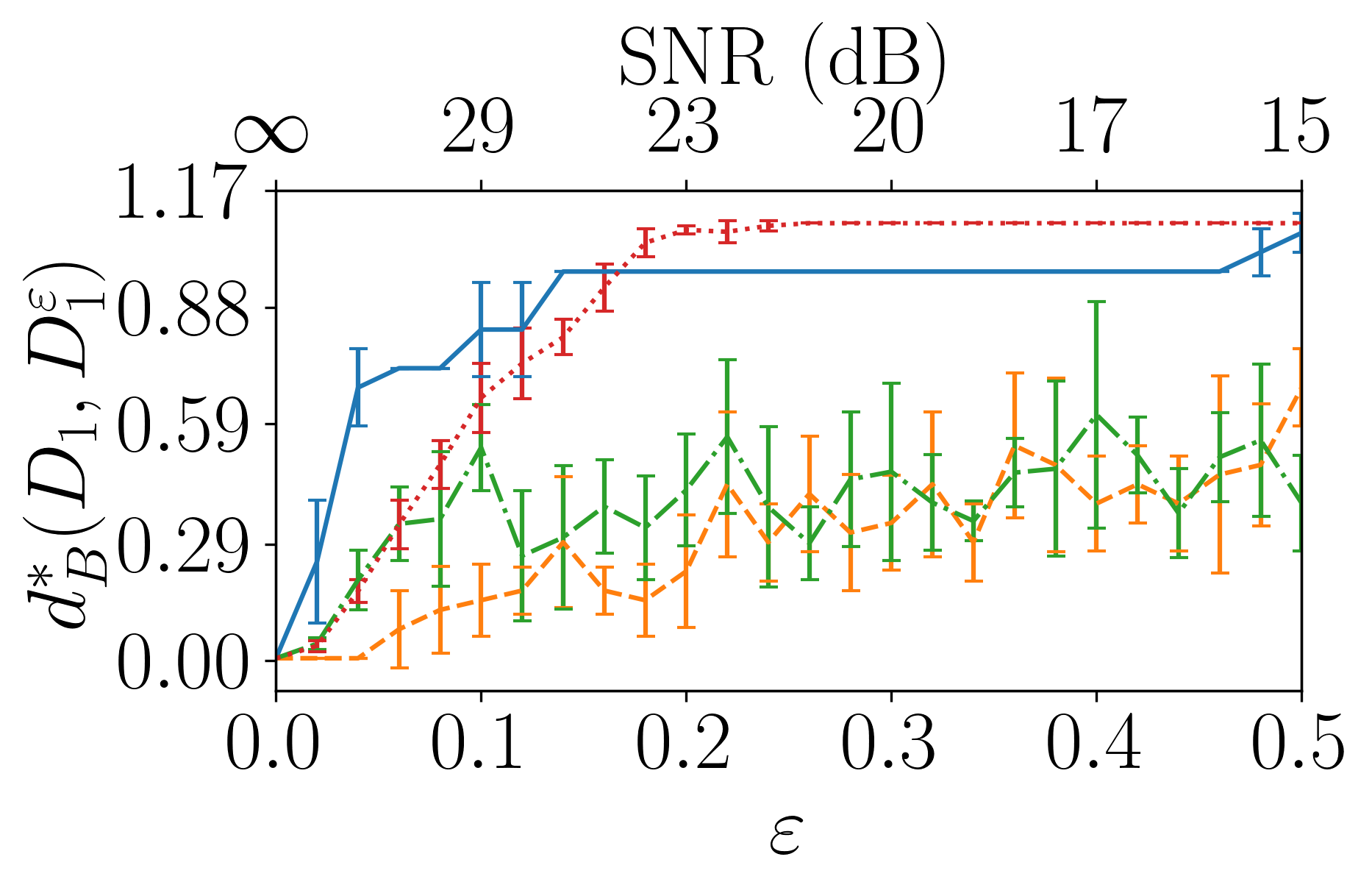}
         {Diffusionless Lorenz Attractor } %
    \end{minipage}
	\hfill
    \begin{minipage}[t]{\myfiguresizeB\textwidth}
        \centering
        \includegraphics[width=\linewidth]{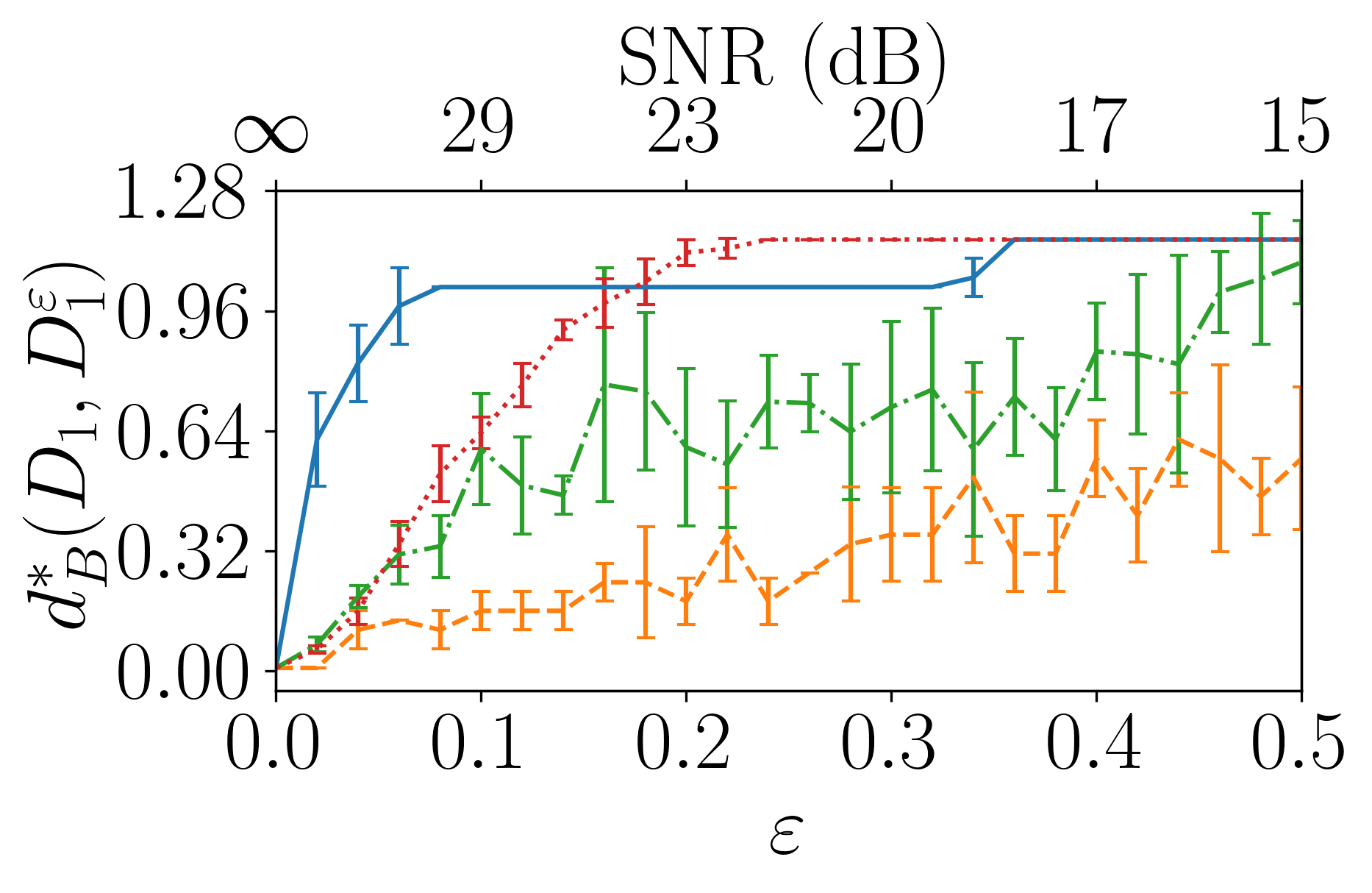}
         {Complex Butterfly Attractor } %
    \end{minipage}
	\hfill
    \begin{minipage}[t]{\myfiguresizeB\textwidth}
        \centering
        \includegraphics[width=\linewidth]{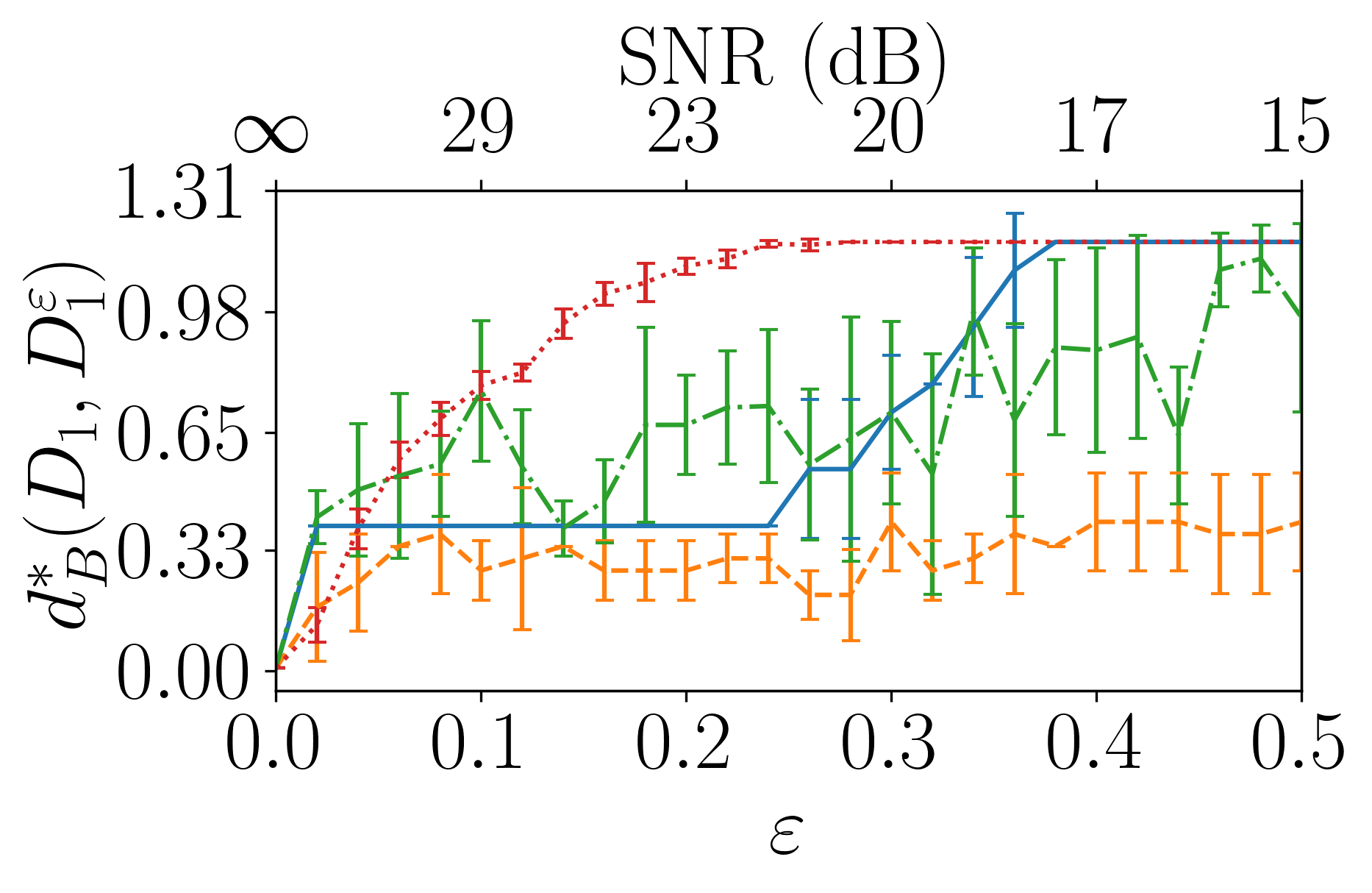}
         {Chen's system } %
    \end{minipage}

    	\vspace{0.2cm}

    \centering
    \begin{minipage}[t]{\myfiguresizeB\textwidth}
        \centering
        \includegraphics[width=\linewidth]{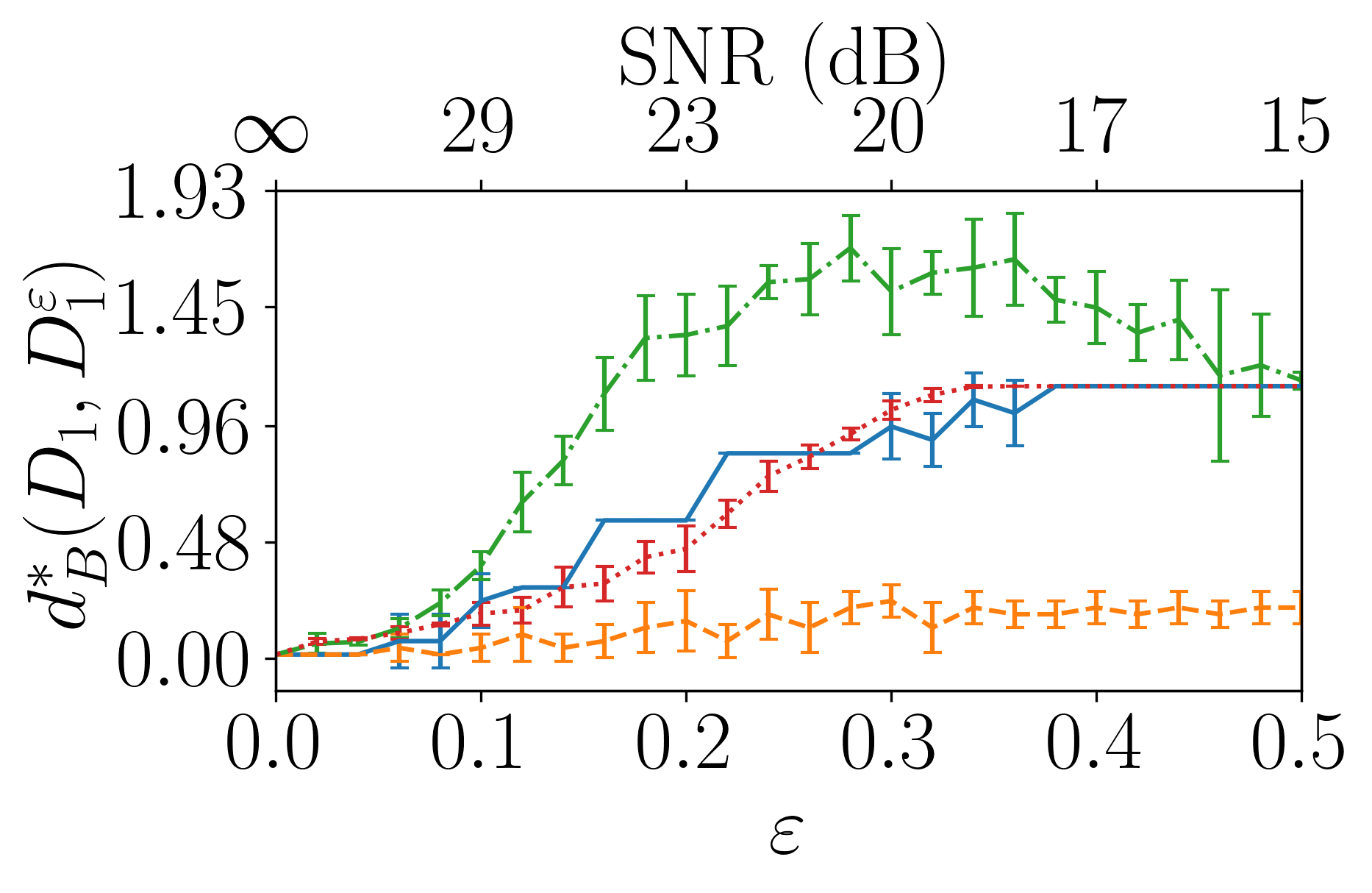}
         {Moore Spiegel Oscillator } %
    \end{minipage}
	\hfill
    \begin{minipage}[t]{\myfiguresizeB\textwidth}
        \centering
        \includegraphics[width=\linewidth]{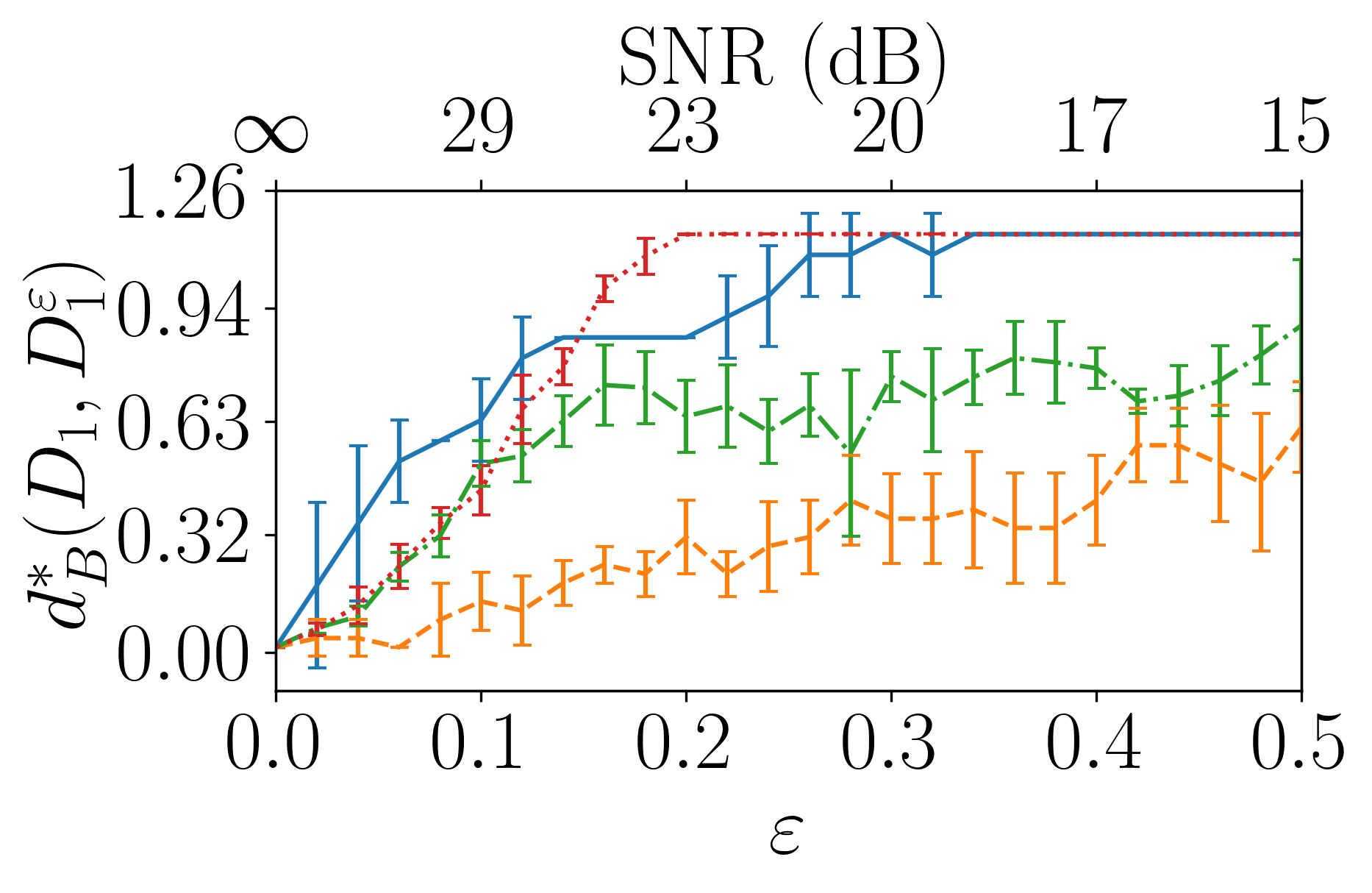}
         {Linear Feedback Rigid Body Motion System } %
    \end{minipage}
	\hfill
    \begin{minipage}[t]{\myfiguresizeB\textwidth}
        \centering
        \includegraphics[width=\linewidth]{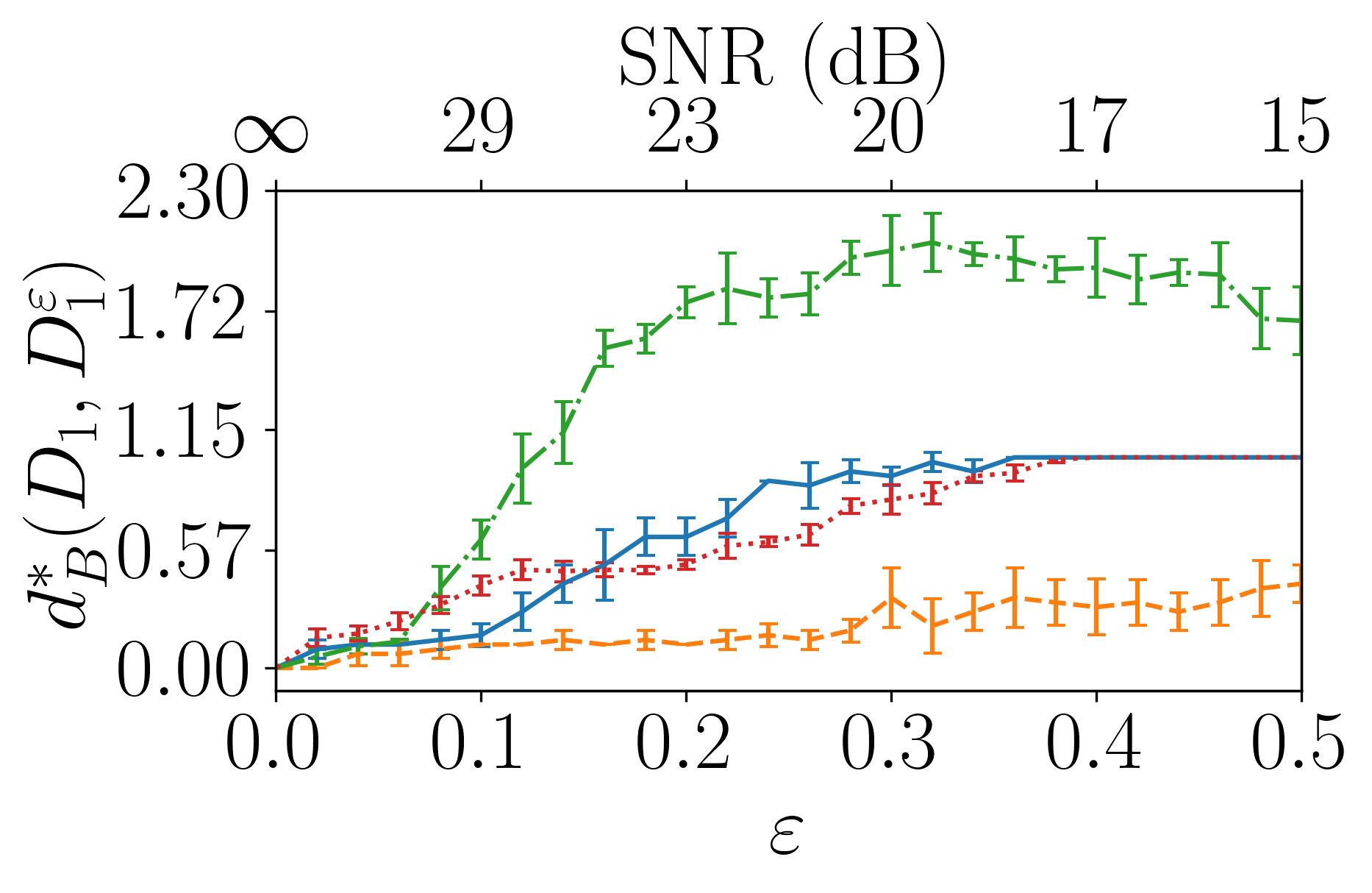}
         Thomas Cyclically Symmetric Attractor  %

    \end{minipage}

    	\vspace{0.2cm}

    \begin{minipage}[t]{\myfiguresizeB\textwidth}
        \centering
        \includegraphics[width=\linewidth]{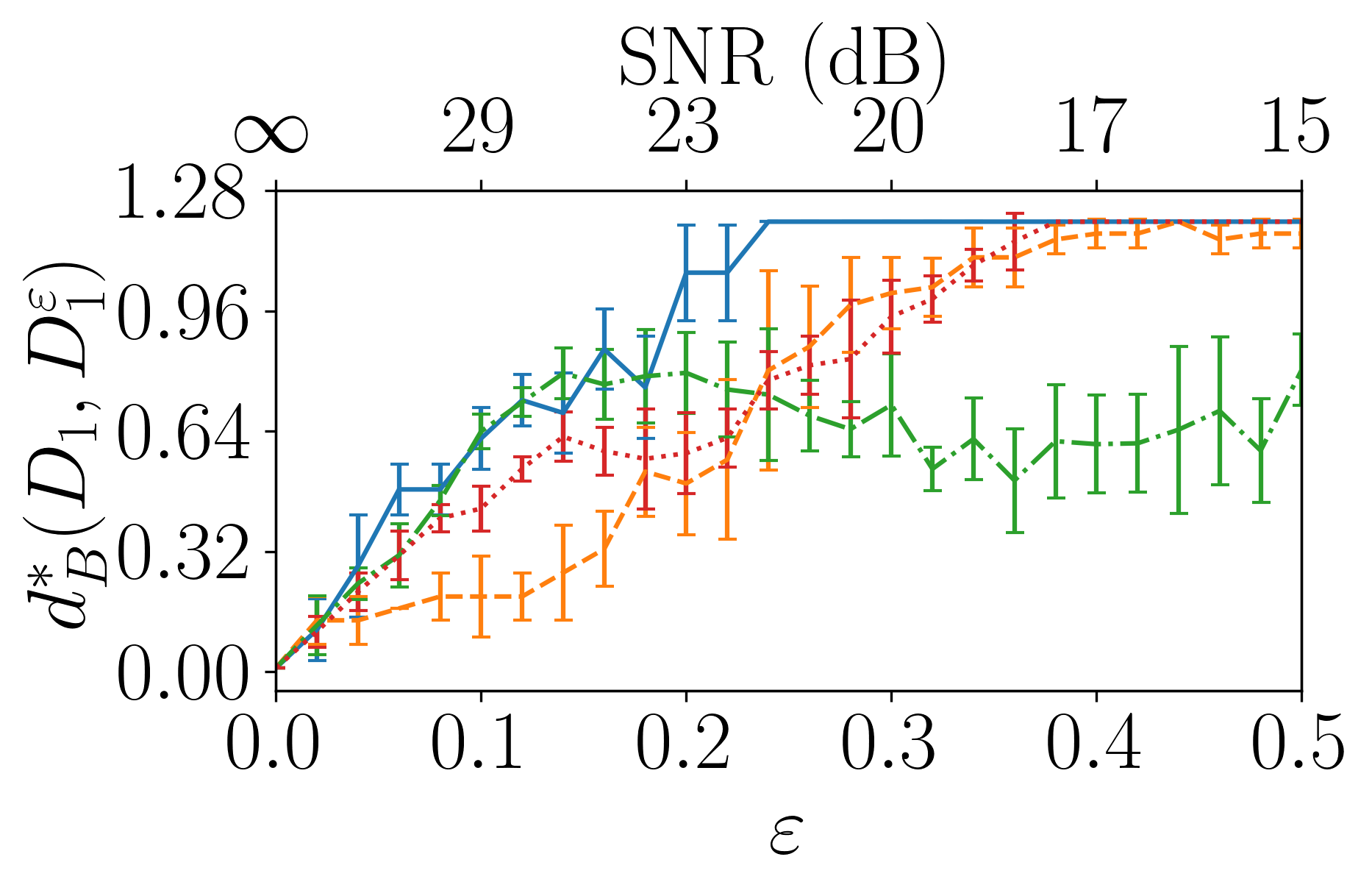}
         {Halvorsens Cyclically Symmetric Attractor } %
    \end{minipage}
	\hfill
    \begin{minipage}[t]{\myfiguresizeB\textwidth}
        \centering
        \includegraphics[width=\linewidth]{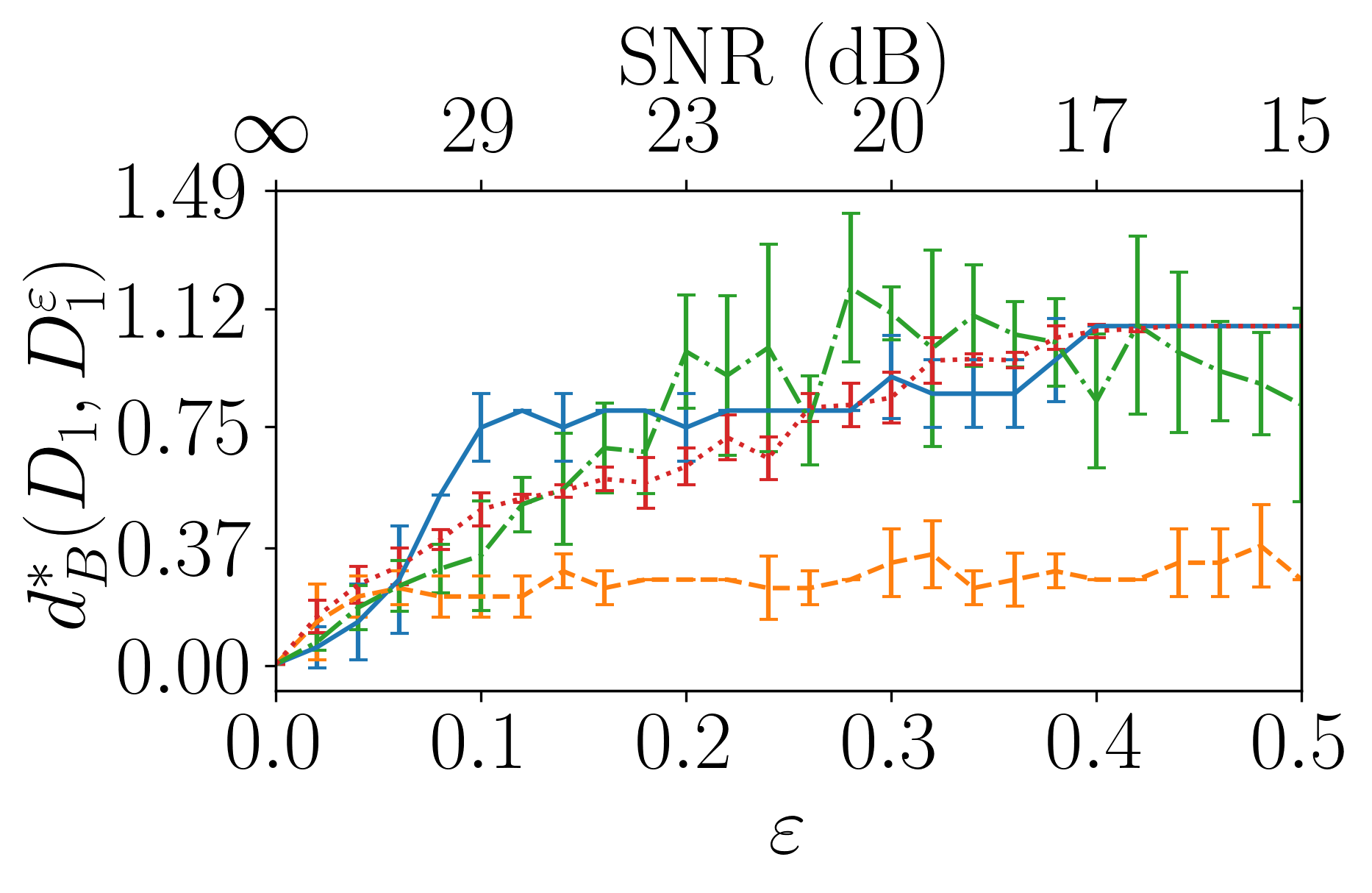}
         ACT Attractor %
    \end{minipage}
	\hfill
    \begin{minipage}[t]{\myfiguresizeB\textwidth}
        \centering
        \includegraphics[width=\linewidth]{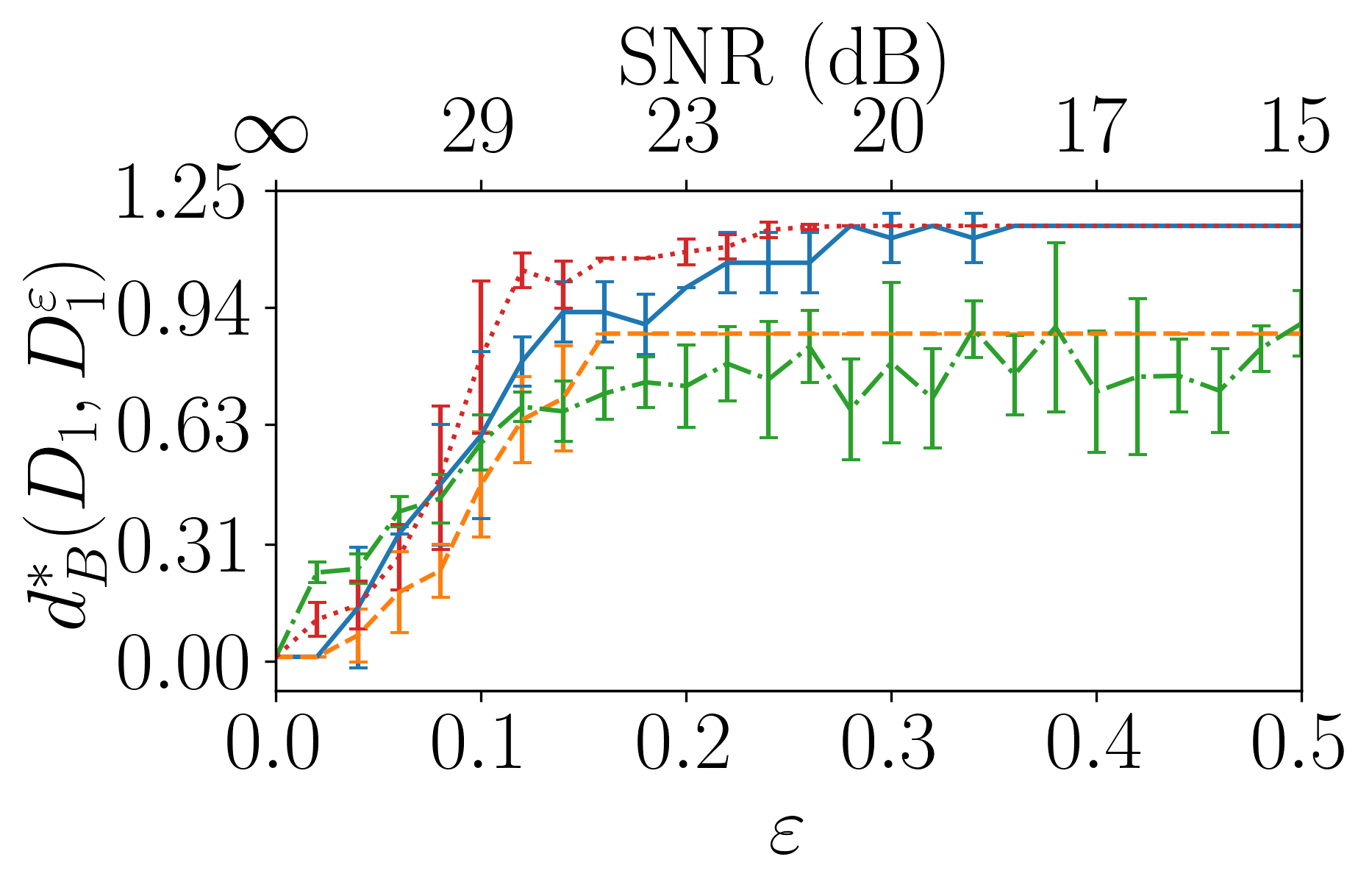}
         {Rabinovich Frabrikant Attractor } %
    \end{minipage}

    \caption{Bottleneck distance stability analysis to standard deviation normalized signal with bounded ($\epsilon = 6\sigma$) Gaussian additive noise.}
\end{figure}

\begin{figure}[h]
    \centering
    \begin{minipage}[t]{\myfiguresizeB\textwidth}
        \centering
        \includegraphics[width=\linewidth]{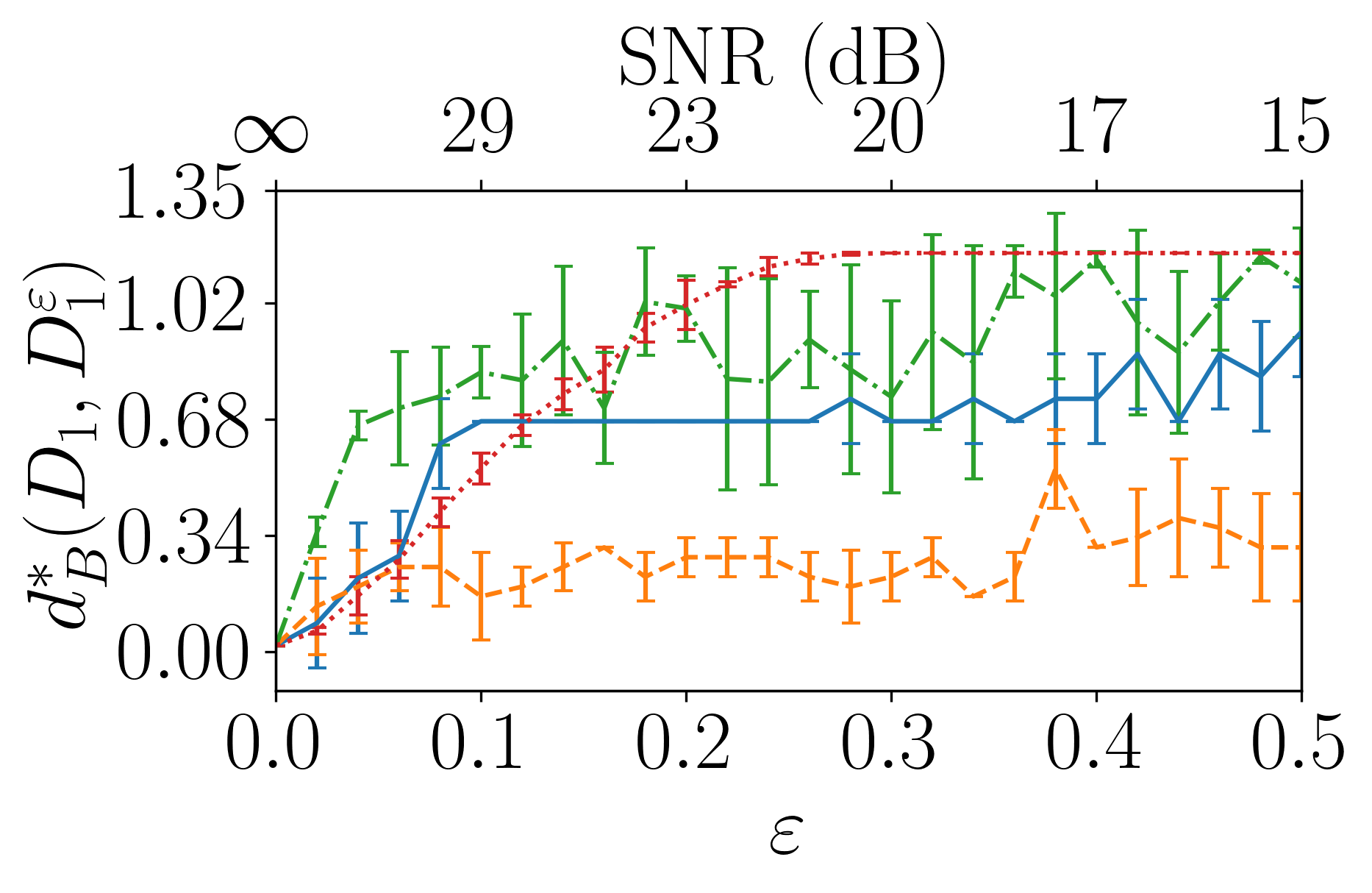}
         {Burke Shaw Attractor } %
    \end{minipage}
	\hfill
    \begin{minipage}[t]{\myfiguresizeB\textwidth}
        \centering
        \includegraphics[width=\linewidth]{noise_stability_periodic_halvorsens_cyclically_symmetric_attractor_all_distances}
         {Halvorsens Cyclically Symmetric Attractor } %
    \end{minipage}
	\hfill
    \begin{minipage}[t]{\myfiguresizeB\textwidth}
        \centering
        \includegraphics[width=\linewidth]{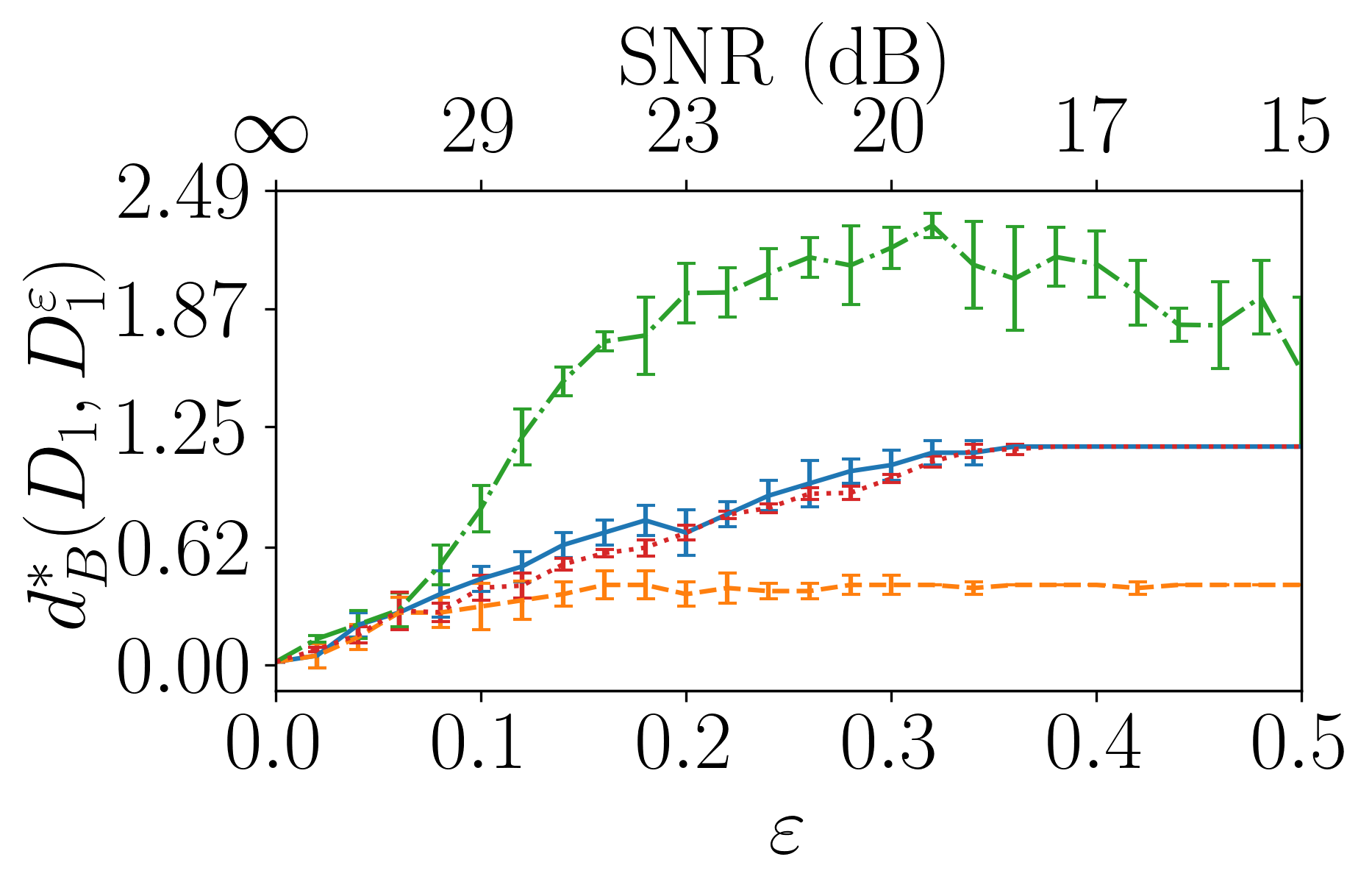}
         {WINDMI } %
    \end{minipage}

    	\vspace{0.2cm}

    \begin{minipage}[t]{\myfiguresizeB\textwidth}
        \centering
        \includegraphics[width=\linewidth]{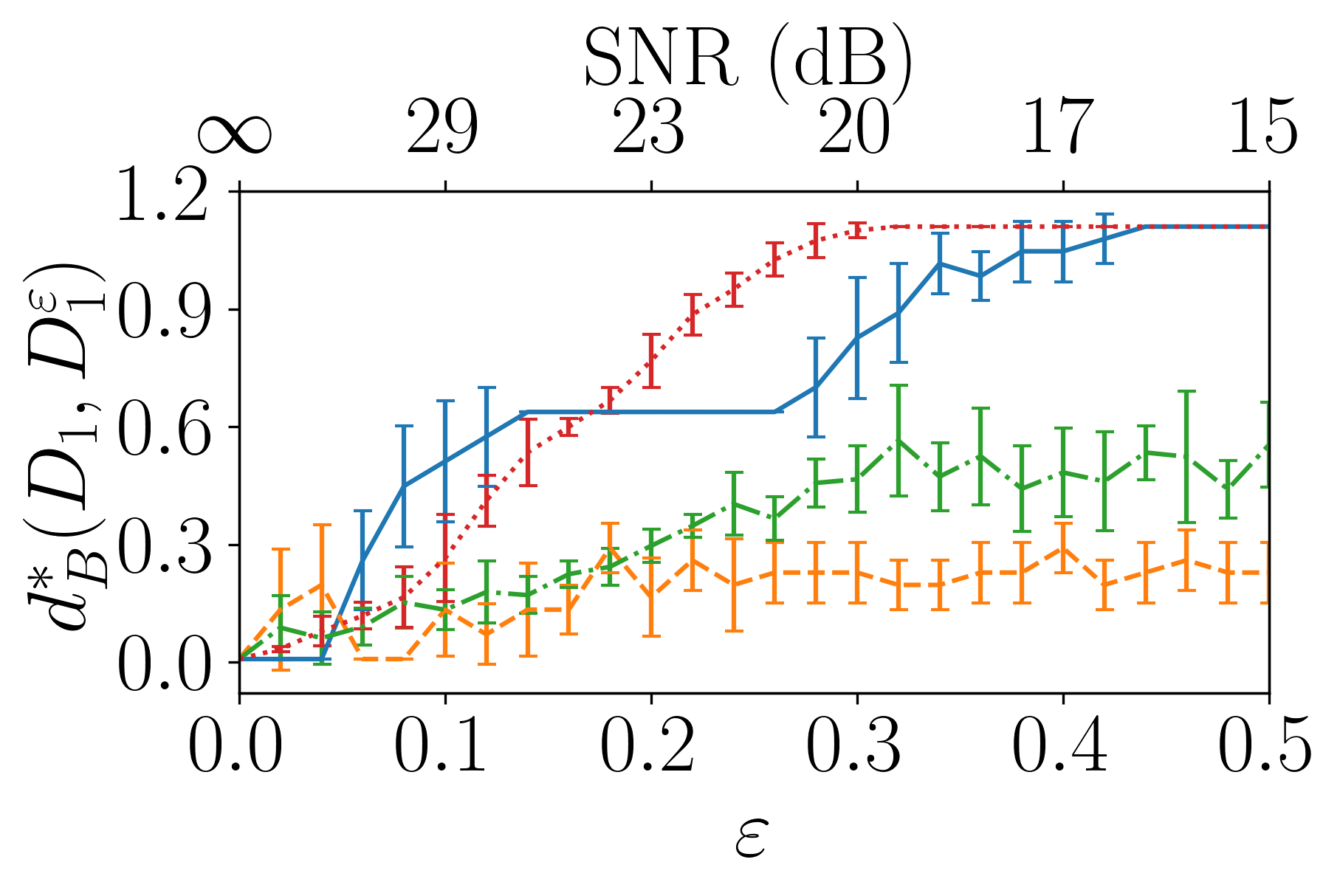}
         {Simplest Cubic Chaotic Flow } %
    \end{minipage}
	\hfill
    \begin{minipage}[t]{\myfiguresizeB\textwidth}
        \centering
        \includegraphics[width=\linewidth]{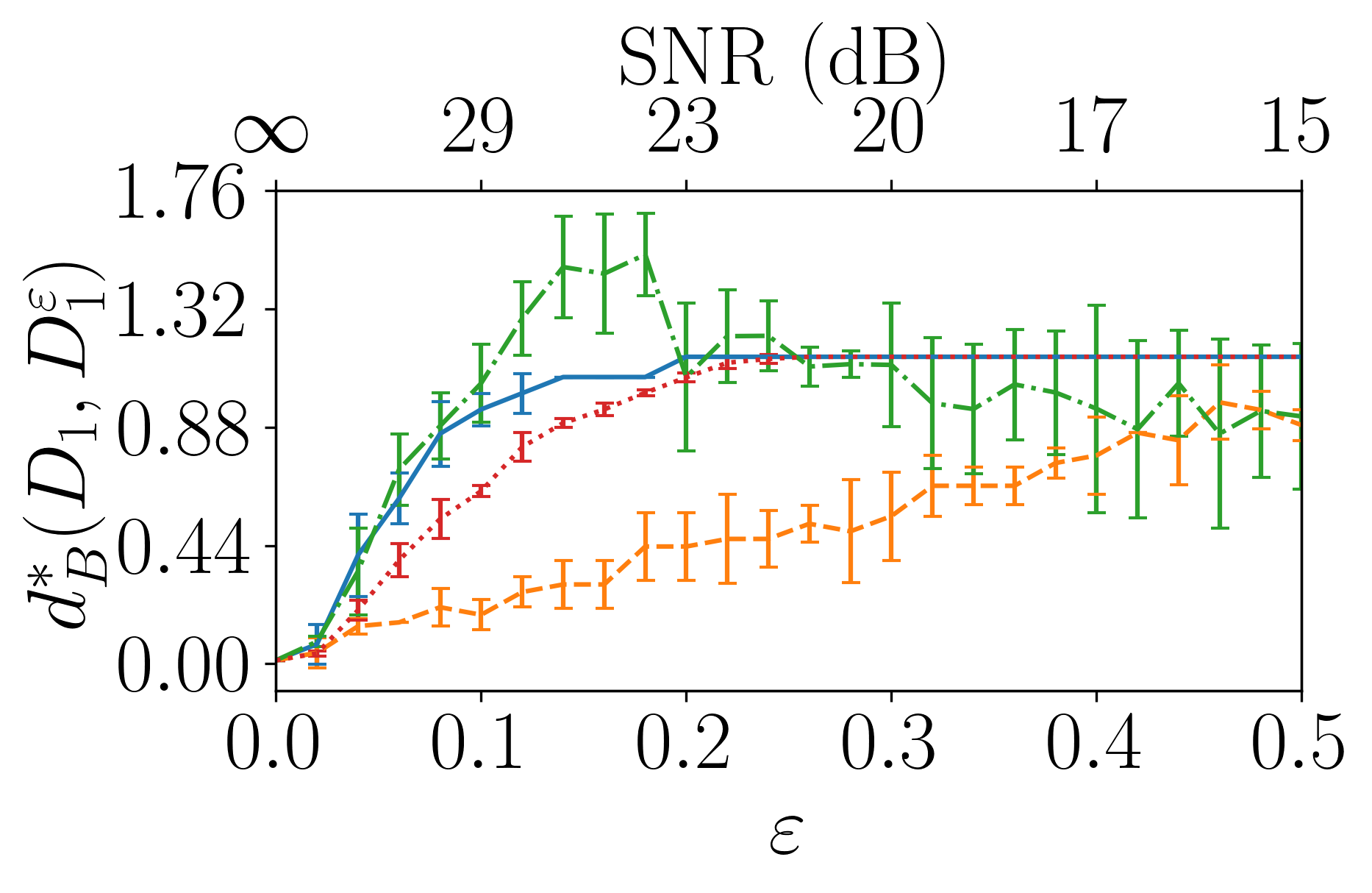}
         {Rucklidge Attractor } %
    \end{minipage}
	\hfill
    \begin{minipage}[t]{\myfiguresizeB\textwidth}
        \centering
        \includegraphics[width=\linewidth]{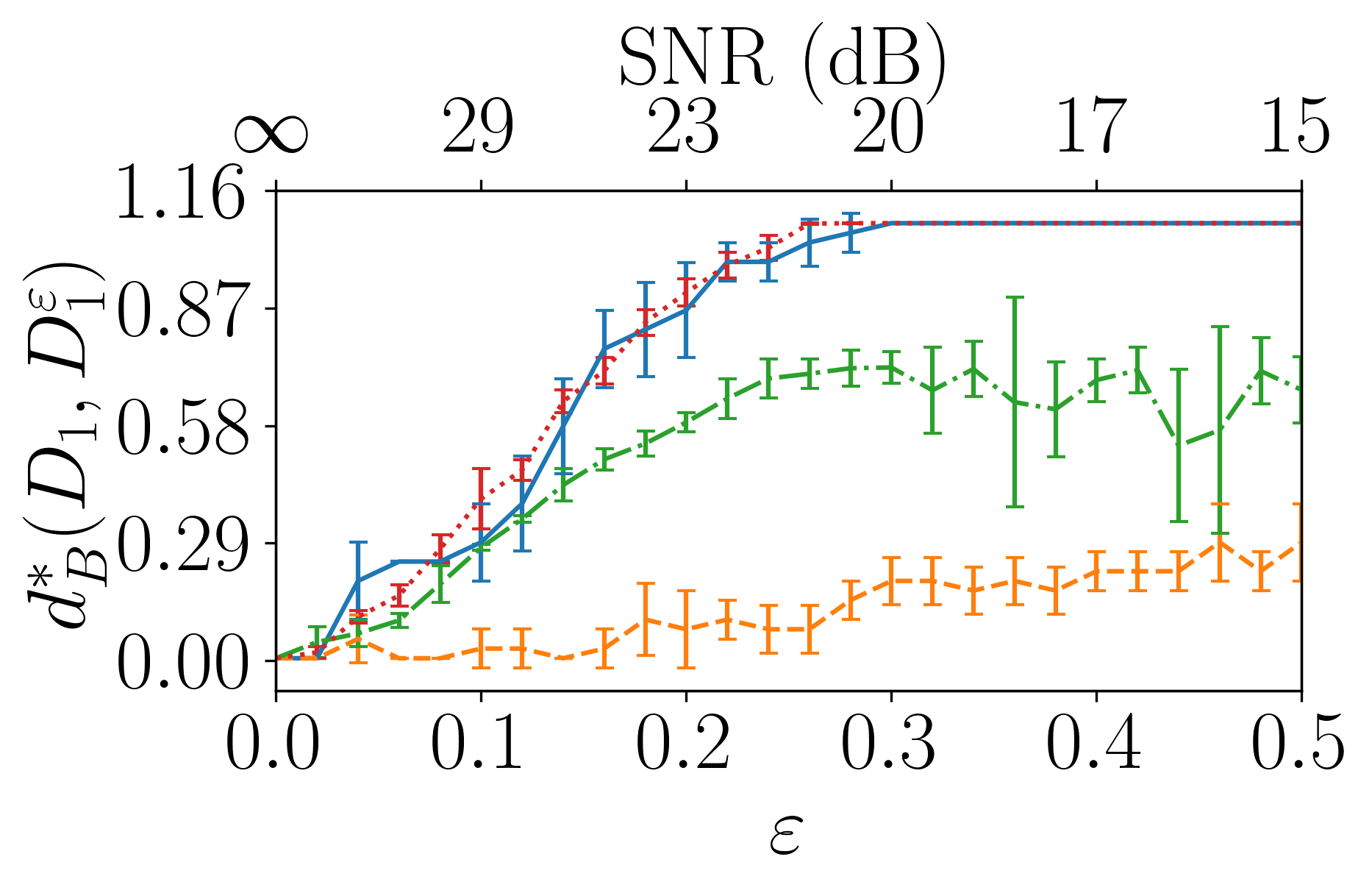}
         {Driven Van Der Pol Oscillator } %
    \end{minipage}

	\vspace{0.2cm}

    \centering
    \begin{minipage}[t]{\myfiguresizeB\textwidth}
        \centering
        \includegraphics[width=\linewidth]{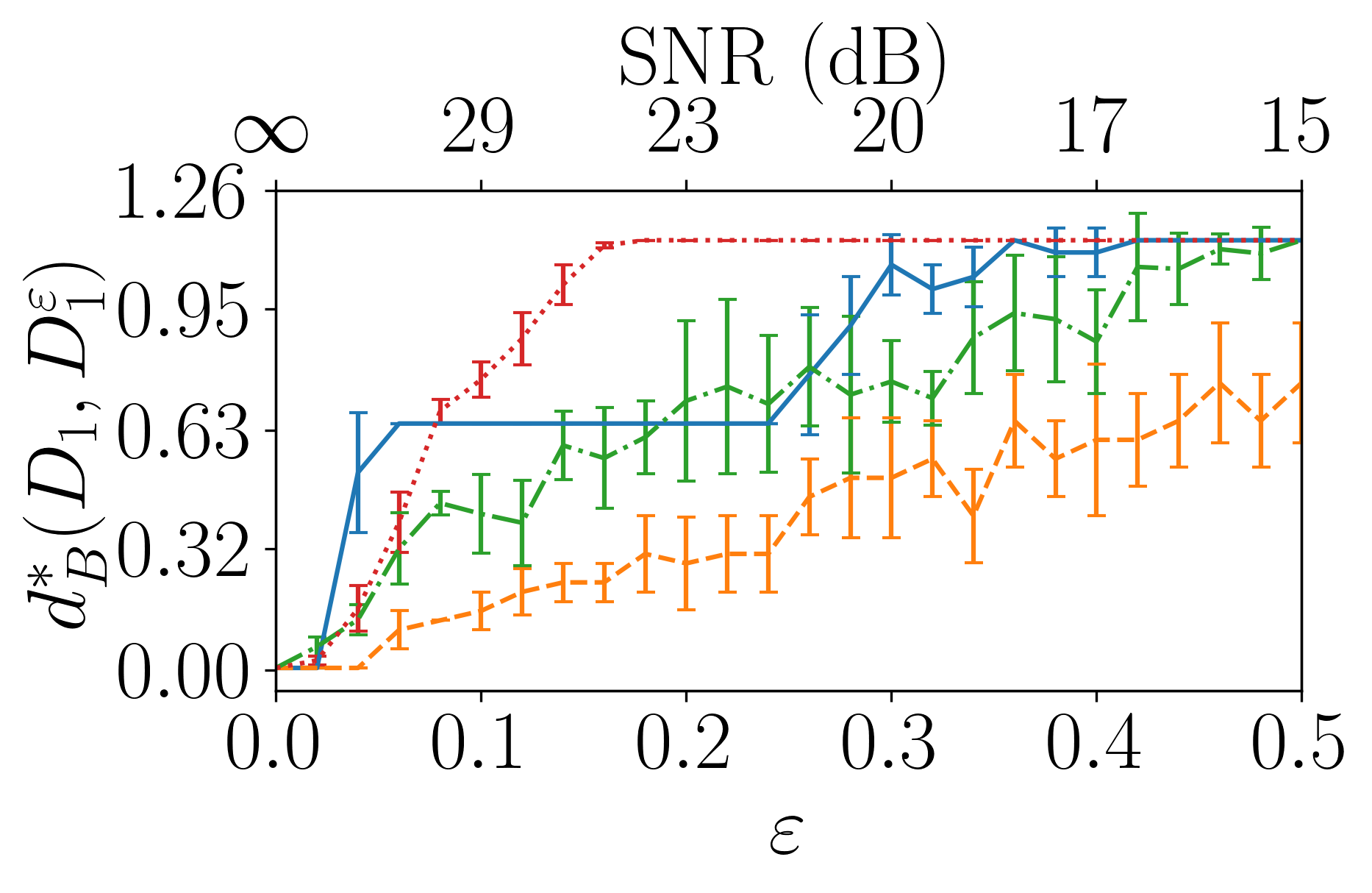}
         {Shaw Van Der Pol Oscillator } %
    \end{minipage}
	\hfill
    \begin{minipage}[t]{\myfiguresizeB\textwidth}
        \centering
        \includegraphics[width=\linewidth]{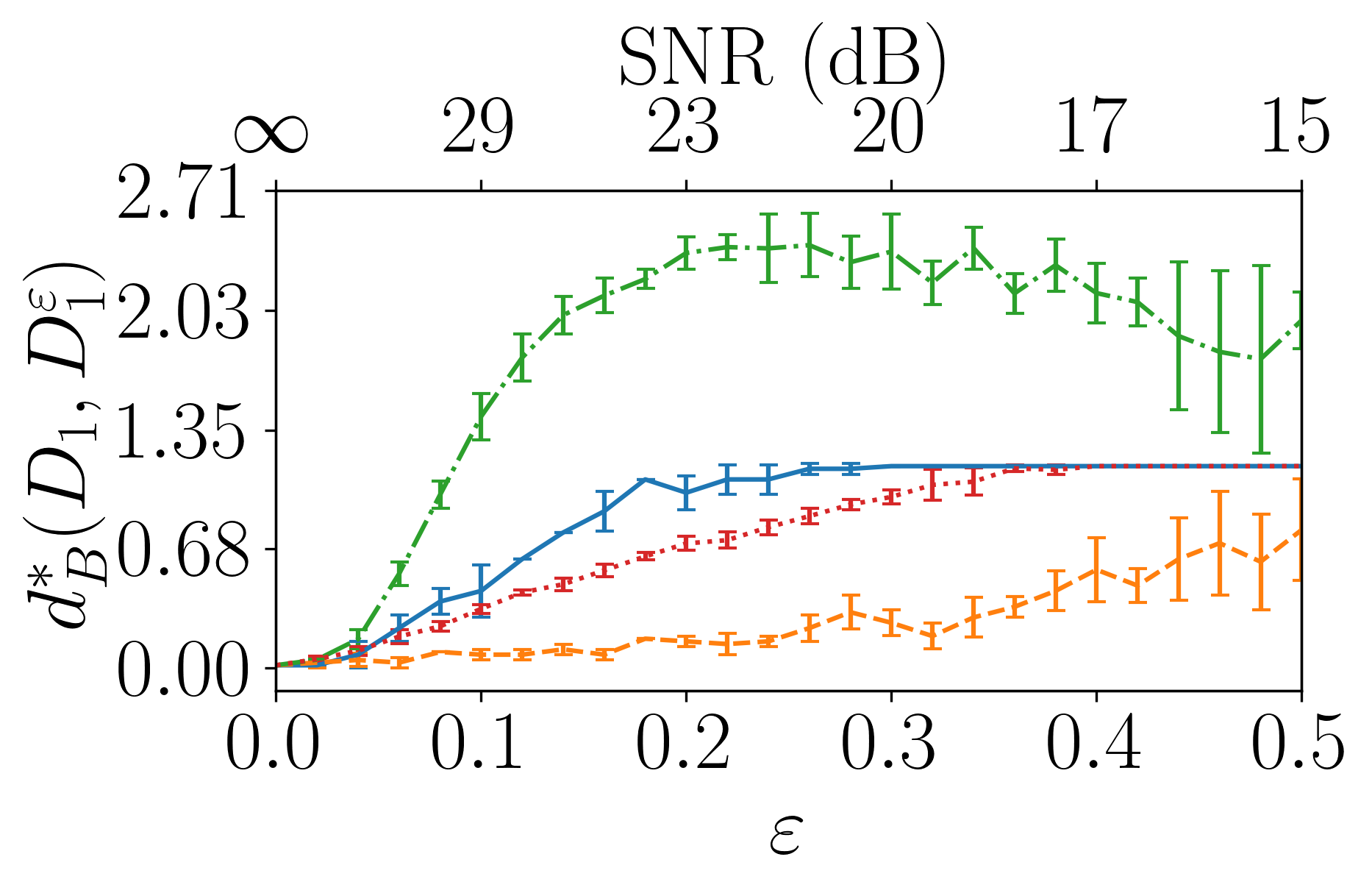}
         {Forced Brusselator } %
    \end{minipage}
	\hfill
    \begin{minipage}[t]{\myfiguresizeB\textwidth}
        \centering
        \includegraphics[width=\linewidth]{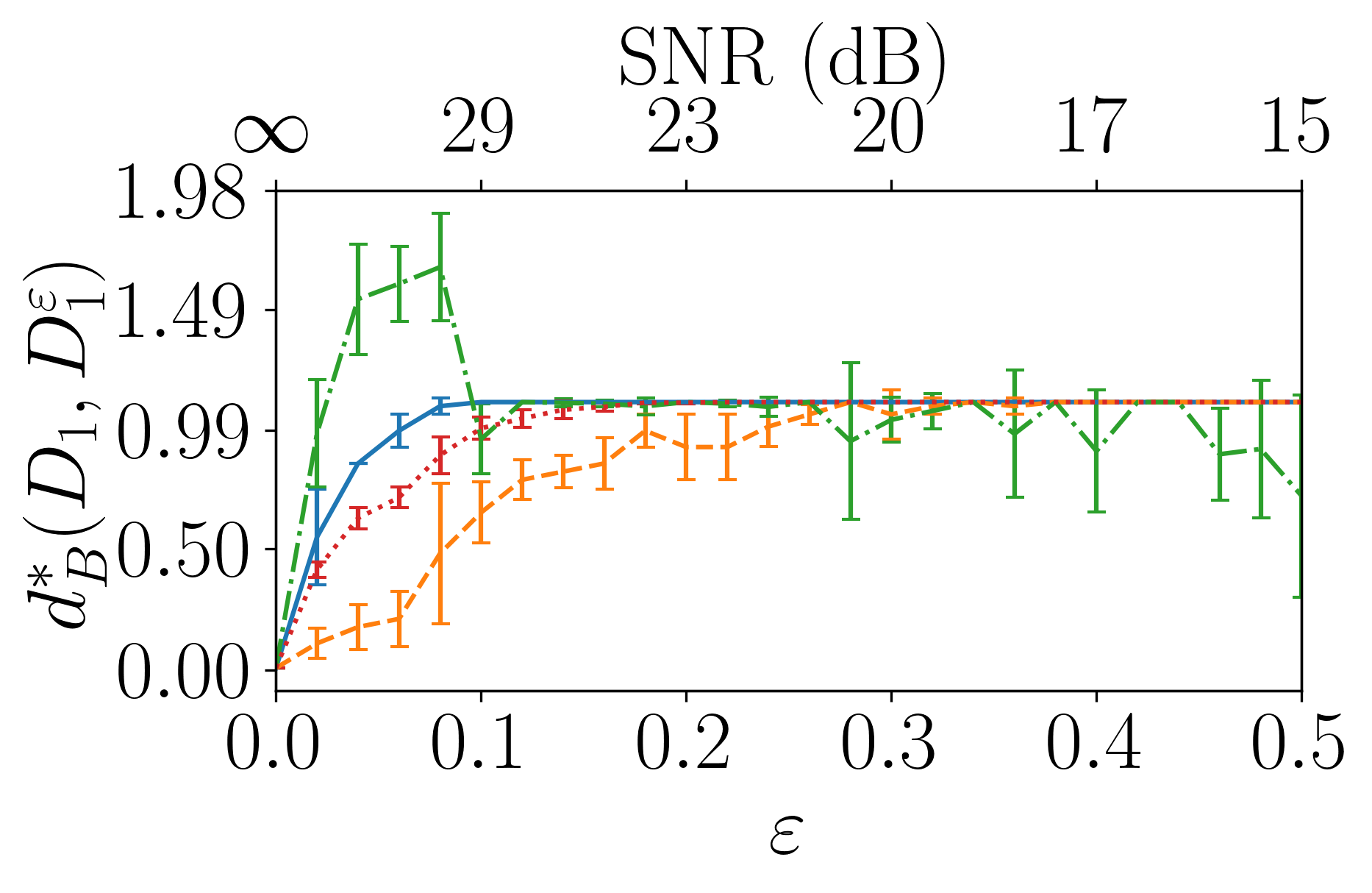}
         {Ueda Oscillator } %
    \end{minipage}

    	\vspace{0.2cm}

    \begin{minipage}[t]{\myfiguresizeB\textwidth}
        \centering
        \includegraphics[width=\linewidth]{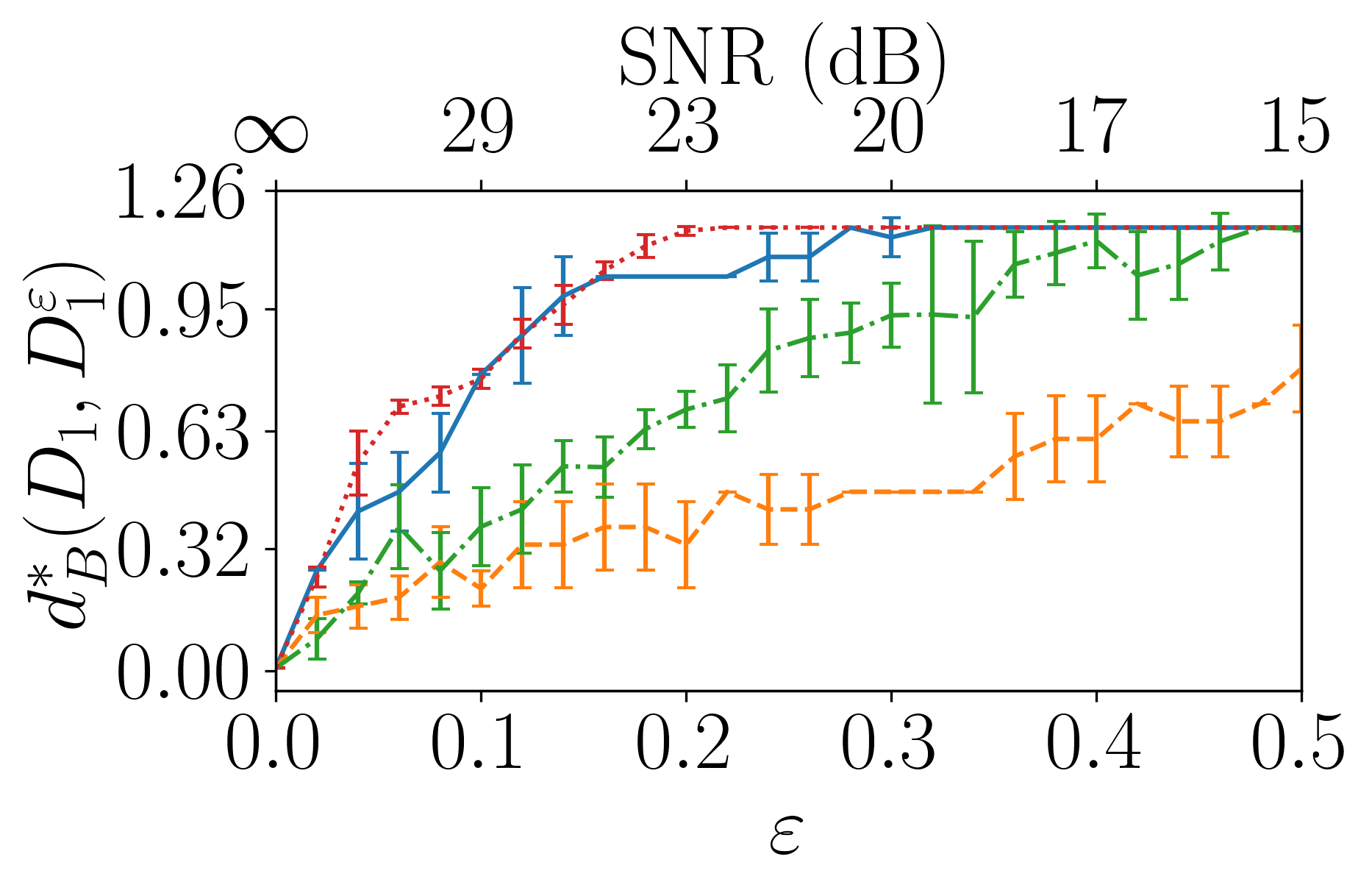}
         {Duffing Van Der Pol Oscillator } %
    \end{minipage}
	\hfill
    \begin{minipage}[t]{\myfiguresizeB\textwidth}
        \centering
        \includegraphics[width=\linewidth]{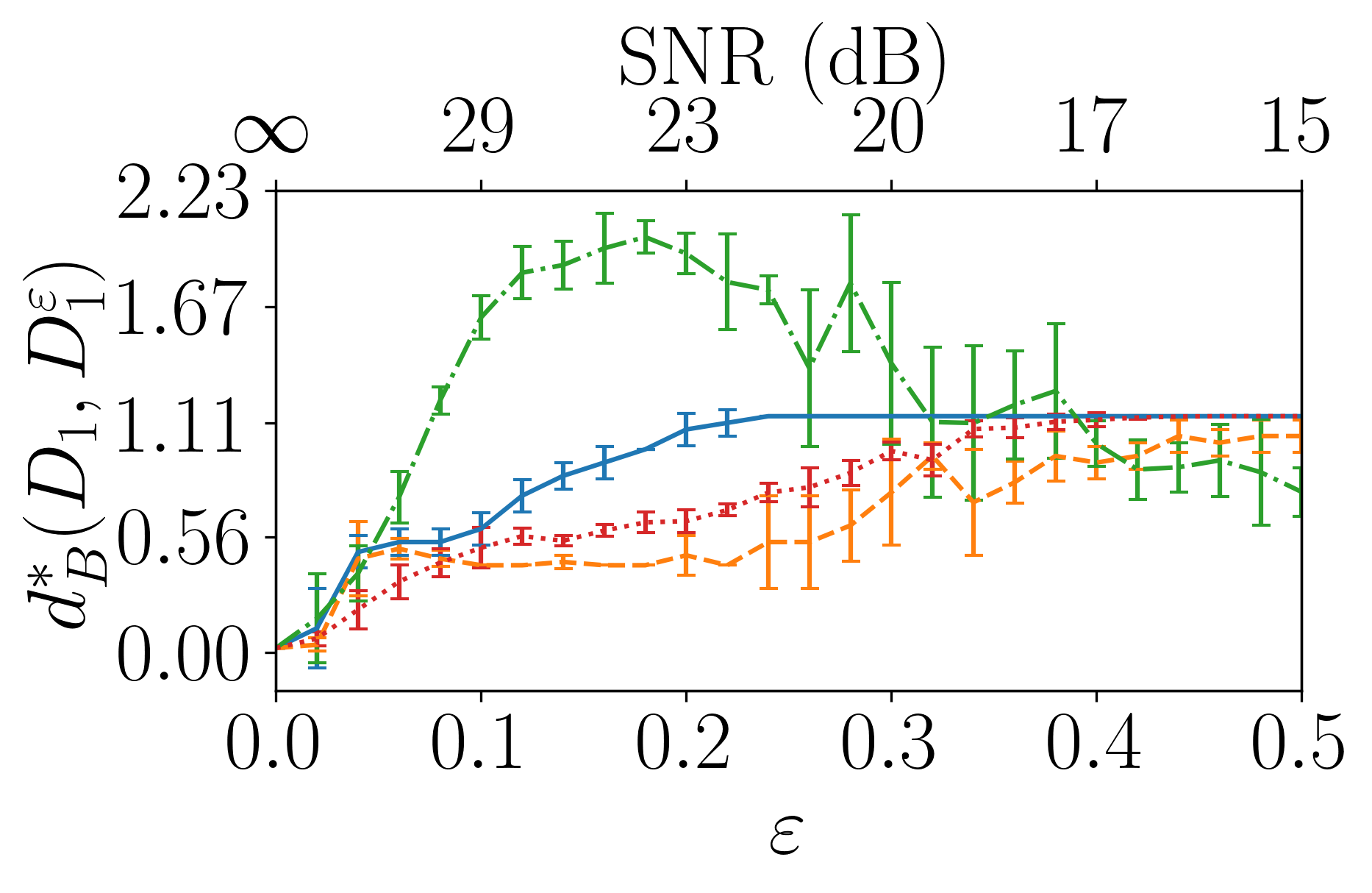}
         {Base Excited Magnetic Pendulum } %
    \end{minipage}

    \caption{Bottleneck distance stability analysis to standard deviation normalized signal with bounded ($\epsilon = 6\sigma$) Gaussian additive noise.}
\end{figure}

\end{document}